\documentclass[a4paper,journal]{IEEEtran}
\usepackage{microtype}
\usepackage{graphicx}
\usepackage[dvipsnames]{xcolor}
\usepackage{subfigure}
\usepackage{booktabs} % for professional tables
\usepackage{hyperref}
\usepackage{bbm}
\usepackage{url}
\usepackage{amsmath}
\usepackage{placeins}% used for \FloatBarrier
\usepackage{amssymb}% used for DeclareMathOperator
\usepackage{cite}
\usepackage{amsmath,amssymb,amsthm}

\theoremstyle{definition}
\newtheorem{example}{Example}

\newcommand{\setnotation}[1]{\mathcal{#1}}

\newcommand{\kld}[2]{D_{\text{KL}}({#1} \Vert {#2})}

\newcommand{\ent}[1]{H(#1)}
\newcommand{\mutinf}[1]{I(#1)}
\newcommand{\indicator}[1]{\mathbbm{1}[#1]}
\newcommand{\stat}[3]{$\text{MG@}{#1}=#2$, $p=#3$}

\begin{document}
\title{Understanding Neural Networks and Individual Neuron Importance via Information-Ordered Cumulative Ablation}

\author{Rana Ali Amjad,~\IEEEmembership{Student Member,~IEEE,} %
Kairen Liu, %
and Bernhard C. Geiger,~\IEEEmembership{Senior Member,~IEEE}%
\thanks{This work was supported by the German Federal Ministry of Education and Research in the framework of the Alexander von Humboldt-Professorship. The work of Bernhard C. Geiger has been funded by the Erwin Schr\"odinger Fellowship J 3765 of the Austrian Science Fund and the iDev40 project. The iDev40 project has received funding from the ECSEL Joint Undertaking (JU) under grant agreement No 783163. The JU receives support from the European Union’s Horizon 2020 research and innovation programme. It is co-funded by the consortium members, grants from Austria, Germany, Belgium, Italy, Spain and Romania. The information and results set out in this publication are those of the authors and do not necessarily reflect the opinion of the ECSEL Joint Undertaking. The Know-Center is funded within the Austrian COMET Program - Competence Centers for Excellent Technologies - under the auspices of the Austrian Federal Ministry for Climate Action, Environment, Energy, Mobility, Innovation and Technology, the Austrian Federal Ministry of Digital and Economic Affairs, and by the State of Styria. COMET is managed by the Austrian Research Promotion Agency FFG.}
\thanks{Rana Ali Amjad was with the Institute for Communications Engineering, Technical University of Munich, Germany. He is now with Amazon, Palo Alto CA, USA. Email: ranaali.amjad@tum.de}
\thanks{Kairen Liu is with Huawei Germany, D\"usseldorf, Germany.}
\thanks{Bernhard C. Geiger is with Know-Center GmbH, Graz, Austria. Email: geiger@ieee.org}
}

\maketitle      

\begin{abstract}
In this work, we investigate the use of three information-theoretic quantities -- entropy, mutual information with the class variable, and a class selectivity measure based on Kullback-Leibler divergence -- to understand and study the behavior of already trained fully-connected feed-forward neural networks. We analyze the connection between these information-theoretic quantities and classification performance on the test set by cumulatively ablating neurons in networks trained on MNIST, FashionMNIST, and CIFAR-10. Our results parallel those recently published by Morcos et al., indicating that class selectivity is not a good indicator for classification performance. However, looking at individual layers separately, both mutual information and class selectivity are positively correlated with classification performance, at least for networks with ReLU activation functions. We provide explanations for this phenomenon and conclude that it is ill-advised to compare the proposed information-theoretic quantities across layers. Furthermore, we show that cumulative ablation of neurons with ascending or descending information-theoretic quantities can be used to formulate hypotheses regarding the joint behavior of multiple neurons, such as redundancy and synergy, with comparably low computational cost. We also draw connections to the information bottleneck theory for neural networks.
\end{abstract}

\section{Introduction}
\label{sec:introduction}
Recent years have seen an increased effort in explaining the success of deep neural networks (NNs) along the lines of several, sometimes controversial, hypotheses. One of these hypotheses suggests that NNs with good generalization performance do not rely on single directions, i.e., the removal of individual neurons has little effect on classification error, and that highly class-selective neurons (especially in shallow layers) may even harm generalization performance~\cite{Deepmind_Cats}.

The claim that class-selectivity is a poor indicator for classification performance has been questioned since its introduction (see Section~\ref{sec:related}). For example,~\cite{Zhou_RevisitingCNNAblation} grants that the effect of class-selective neurons on overall classification performance is minor, but shows that it can be very large for individual classes. The authors of~\cite{Meyes_Ablation} showed that some neuron outputs represent features relevant for multiple classes while others represent class-specific features, indicated that ablating both types of neurons can lead to drops in overall classification performance, that certain layers appear to be particularly important for certain classes, and that some layers exhibit redundancy. Finally,~\cite{Ukita_Orientation} considered feature- rather than class-selectivity and showed that ablating highly orientation-selective neurons from shallow layers harms classification performance. The interplay between individual neurons, redundancy, class specificity, and classification performance thus seems to be more intricate than expected.

As our first main contribution, our work complements~\cite{Zhou_RevisitingCNNAblation,Meyes_Ablation,Ukita_Orientation} and provides yet another perspective on the results of~\cite{Deepmind_Cats}. We propose information-theoretic quantities to measure the importance of individual neurons for the classification task. Specifically, we investigate how the variability, class selectivity, and class information of a neuron output (Section~\ref{sec:impfunc}) connect with classification performance when said neuron is ablated (Section~\ref{sec:understanding}). Our experiments rely on fully-connected feed-forward NNs trained on the MNIST, FashionMNIST, and CIFAR-10 datasets, as these datasets have evolved into benchmarks for which the results are easy to understand intuitively. In Section~\ref{subsec:pruning}, we show that neither class selectivity nor class information are good performance indicators when ablating neurons across layers, thus confirming the results in~\cite{Deepmind_Cats}. However, we observe 1) that class information and class selectivity values differ greatly from layer to layer (Section~\ref{subsec:layerdistribution}) and 2) that for NNs with ReLU activation functions, class information and class selectivity are positively correlated with classification performance when cumulative ablation is performed separately for each layer (Sections~\ref{subsec:layerpruning:MNIST} to~\ref{subsec:layerpruning:CIFAR}). Our results are not in contrast with those in~\cite{Deepmind_Cats}, but complement them and can be reconciled with them with reference to Simpson's paradox. In Section~\ref{sec:discussion}, we briefly discuss the implications of these findings on NN understanding, the design of objective functions for NN training, and neuron pruning, a recent trend for reducing the complexity of NNs~\cite{he2014reshaping,srinivas2015data,Mariet_Divnet,Li_CNNPruning,Molchanov_CNNPruning}. 

As our second main contribution, we show that \emph{information-ordered cumulative ablation} -- where we rank a set of neurons in terms of a specific information-theoretic quantity and then set to zero the activations of a fixed number of neurons based on this ranking -- can suggest hypotheses regarding the behavior of an entire layer. More specifically, we believe that such analyses can help in estimating the degree of synergy of a layer and the number of active and/or redundant neurons; such insights are required to consolidate this work and~\cite{Deepmind_Cats}, which look at individual neurons separately, with the works in the spirit of the information bottleneck principle~\cite{Tishby_BlackBox,Saxe_IBTheory,Amjad_HowNotTo}, which analyze the entire layer. As such, information-ordered cumulative ablation hints at results that can be obtained using \emph{partial information decomposition}~\cite{Williams_PID,Rauh_PID}, a decomposition of the mutual information between the class variable and an entire hidden layer into unique, redundant, and synergistic contributions. While partial information decomposition (PID) requires computing information quantities for exponentially many (in the number of neurons per layer) \emph{subsets} of neurons (thus suffering from the curse of dimensionality and prohibitive computational complexity), information-ordered ablation only requires computing these quantities for all individual neuron outputs in the considered layer. Therefore, our analyses in Sections~\ref{subsec:layerpruning:MNIST} to~\ref{subsec:layerpruning:CIFAR} provide relevant insights into the behavior of NNs at a comparably low computational costs.

\section{Related Work} \label{sec:related}

Morcos et al.~\cite{Deepmind_Cats} studied the dependence of NN classification performance on the output of individual neurons (``single direction'') via ablation analyses. Computing the class selectivity of each neuron, they showed that there is little correlation between this quantity and the performance drop of the NN when said neuron is ablated. They therefore concluded that class selectivity (and the mutual information between the neuron output and the class variable) is a poor indicator of NN performance.

The authors of~\cite{Zhou_RevisitingCNNAblation} provided a different perspective on the results in~\cite{Deepmind_Cats} by showing that, although the effect of ablating individual neurons on overall classification performance is small, the effect on the classification of individual classes can be large. For example, ablating the ten most informative neurons from the fifth convolutional layer of AlexNet for a given class of the Places dataset makes its detection probability drop by more than 40\%, on average. In a similar experiment, the authors of~\cite{Bau_Understanding} showed that by ablating as little as four of the 512 output channels of the last convolutional layer of VGG-16 can cause the detection probability of the ``ski resort'' class drop from 81.4\% to 64.0\%. Similar results along that line were shown in~\cite{Meyes_Ablation}, where the authors performed individual and pairwise ablations in a NN with two hidden layers trained on MNIST. They showed that some neurons encode general features, affecting overall classification accuracy strongly, while some neurons encode class-specific features, the ablation of which has less (but still noticeable) effect on the classification performance. Both~\cite{Meyes_Ablation} and~\cite{Zhou_RevisitingCNNAblation} observed that ablating neurons with class-specific features can have positive effect on the detection of an unrelated class, suggesting implications for targeted weight pruning. Meyes et al.~\cite{Meyes_Ablation} further discovered that pairwise ablation often has a stronger effect than the summed effect of ablating individual neurons, indicating that intermediate layers exhibit redundancy in some cases. Ablating a certain fraction of filters in different layers of a VGG-19 trained on ImageNet showed that different layers have different sensitivity to ablation, and that this sensitivity is also class-dependent, i.e., some classes suffer more from ablating filters in a given layer than others~\cite{Meyes_Ablation}. The authors of~\cite{Dalvi_GrainofSalt} performed ablation analysis in LSTM neural language models and LSTMs trained for machine translation. They ranked neurons via their linguistic importance by training a logistic classifier on neuron outputs and observing the weight the classifier places on a given neuron. They discovered that certain linguistic properties are represented by few neurons, while other properties are highly distributed, and that ablating a fixed fraction of linguistically important neurons harms part-of-speech tagging and semantic tagging tasks more strongly than ablating the same fraction of linguistically unimportant neurons.

That class-selectivity increases towards deeper layers was observed in~\cite{Deepmind_Cats,Yu_CriticalPaths}. More specifically, the authors of~\cite{Yu_CriticalPaths} claim that deeper layers in a CNN are specific for a single class. This allows distilling so-called critical paths by retaining only these class-specific neurons in the deeper layers to obtain a CNN trained in a one-vs-all classification task. In contrast to class-selectivity, orientation-selectivity, a special kind of feature-selectivity in convolutional NNs, appears to occur in layers at different depths~\cite{Ukita_Orientation}. Ablating these orientation-selective filters in shallow layers harms classification performance more than ablating unselective filters, and ablating filters in deeper layers has little overall effect~\cite[Fig.~5]{Ukita_Orientation}.

\section{Setup and Preliminaries}\label{sec:setup}
We consider classification via fully-connected feed-forward NNs, i.e., the task of assigning data sample $x$ to a class in $\setnotation{C}$, $|\setnotation{C}|=C$. We assume that the parameters of the NN had been learned from the labeled training set $\setnotation{D}_t$. We moreover assume that we have access to a labeled validation set that was left out during training. We denote this dataset by $\setnotation{D}:=\{(x_1,y_1),...,(x_N,y_N)\}$, in which $x_i$ is the $i$-th data sample and $y_i$ the corresponding class label. We assume that $N\gg C$.

Let $t_j^{(i)}(x_\ell)$ denote the output of the $j$-th neuron in the $i$-th layer of the NN if $x_\ell$ is the data sample at the input. With $w^{(i-1)}_{p,j}$ denoting the weight connecting the $p$-th neuron in the $(i-1)$-th layer to the $j$-th neuron in the $i$-th layer, $b^{(i)}_j$ denoting the bias term of the $j$-th neuron in the $i$-th layer, and $\sigma{:}\ \mathbb{R}\to\mathbb{R}$ denoting an activation function, we obtain $t_j^{(i)}(x_\ell)$ by setting
\begin{equation}
t_j^{(i)}(x_\ell) = \sigma\left(b^{(i)}_j+\sum_{p} w^{(i-1)}_{p,j} t_p^{(i-1)}(x_\ell) \right)
\end{equation}
and by setting $t_j^{(0)}(x_\ell)$ to the $j$-th coordinate of $x_\ell$. The output of the network is a softmax layer with $C$  neurons, each corresponding to one of the $C$ classes.

We assume that the readers are familiar with information-theoretic quantities such as entropy, mutual information and Kullback-Leibler (KL) divergence, cf.~\cite[Ch.~2]{Cover_Information}.
To be able to use such quantities to measure the importance of individual neurons in the NN, we treat class labels, NN input, and neuron outputs as random variables (RVs). To this end, let $Q{:}\ \mathbb{R}\to\setnotation{T}$ be a quantizer that maps neuron outputs to a finite set $\setnotation{T}$. Now let $Y$ be a RV over the set $\setnotation{C}$ of classes and $T^{(i)}_{j}$ a RV over $\setnotation{T}$, corresponding to the quantized output of the $j$-th neuron in the $i$-th layer. We define the joint distribution of $Y$ and $T^{(i)}_{j}$ via the joint frequencies of $\{(y_\ell,Q(t_j^{(i)}(x_\ell)))\}$ in the validation set, i.e., $\forall c\in\setnotation{C},t\in\setnotation{T}$,
\begin{equation} \label{eq:joint_dist_estimation}
  P_{Y,T_j^{(i)}}(c,t) = \frac{\sum_{\ell=1}^N \indicator{y_\ell=c,Q(t_j^{(i)}(x_\ell))=t} }{N}
\end{equation}
where $\indicator{\cdot}$ is the indicator function. The assumptions that $N\gg C$ and that $|\setnotation{T}|$ is small obviate the need for more sophisticated estimators for the distribution $P_{Y,T_j^{(i)}}$, such as Laplacian smoothing.

\section{Information-Theoretic Quantities for Measuring Individual Neuron Importance}\label{sec:impfunc}
In this section, we propose information-theoretic quantities as candidate importance measures for neurons in a NN; each of these measures can be computed from the validation set $\setnotation{D}$ with a complexity of $\mathcal{O}(N)$~\cite[Appendix~F.7]{Amjad_Understanding_arXiv}. 

\textbf{Entropy.}  
Entropy quantifies the uncertainty of a RV. In the context of a NN, the entropy
\begin{equation}\label{eq:ent}
 \ent{T^{(i)}_{j}} = -\sum_{t\in\setnotation{T}} P_{T_j^{(i)}}(t) \log P_{T_j^{(i)}}(t)
\end{equation}
indicates if the neuron output varies enough to fall into different quantization bins for different data samples. It has been proposed as an importance function for pruning in~\cite{he2014reshaping}. 

\textbf{Mutual Information.}
While zero entropy of a neuron output suggests that it has little influence on classification performance (see also Example~\ref{ex:entropy} below), the converse is not true, i.e., high entropy does not imply that the neuron is important for classification. Indeed, the neuron may capture a highly varying feature of the input that is irrelevant for classification. As a remedy, the mutual information between the neuron output and the class variable, i.e.,
\begin{equation}\label{eq:MI}
 \mutinf{T^{(i)}_{j};Y} = \ent{T^{(i)}_{j}}-\ent{T^{(i)}_{j}|Y}
\end{equation}
measures how the knowledge of $T^{(i)}_{j}$ helps predicting $Y$. It was used to characterize neuron importance in, e.g.,~\cite{Deepmind_Cats}, and appears in corresponding classification error bounds~\cite{Verdu_GeneralizingFano}. It can be shown that neurons with small $\ent{T^{(i)}_{j}}$ also have small $\mutinf{T^{(i)}_{j};Y}$, cf.~\cite[Th.~2.6.5]{Cover_Information}.

\textbf{Kullback-Leibler Selectivity.}
It has been observed that, especially at deeper layers, the activity of individual neurons may distinguish one class from all others. Mathematically, for such a neuron there exists a class $y$ such that the class-conditional distribution $P_{T_j^{(i)}|Y=y}$ differs significantly from the marginal distribution $P_{T_j^{(i)}}$, i.e., the \emph{specific information} (cf.~\cite{Williams_PID}) $\kld{P_{T_j^{(i)}|Y=y}}{P_{T_j^{(i)}}}$ is large. Neurons with large specific information for at least one class may be useful for classification, but may nevertheless be characterized by low entropy $\displaystyle\ent{T^{(i)}_{j}}$ and low mutual information $\mutinf{T^{(i)}_{j};Y}$, especially if the number $C$ of classes is large. We therefore propose the maximum specific information over all classes as a measure of neuron importance:
\begin{equation}\label{eq:KLDclassmax}
 \max_{y\in \setnotation{C}} \quad \kld{P_{T_j^{(i)}|Y=y}}{P_{T_j^{(i)}}}
\end{equation}
This quantity is high for neurons that activate differently for exactly one class and can thus be seen as an information-theoretic counterpart of the selectivity measure used in~\cite{Deepmind_Cats}. We therefore call the quantity defined in~\eqref{eq:KLDclassmax} \emph{Kullback-Leibler (KL) selectivity}. Specifically, KL selectivity is maximized if all data samples of a specific class label are mapped to one value of $T^{(i)}_{j}$ and all the other data samples (corresponding to other class labels) are mapped to other values of $T^{(i)}_{j}$. In this case, $T^{(i)}_{j}$ can be used to distinguish this class label from the rest. KL selectivity is an upper bound on mutual information and is zero if and only if the mutual information $\mutinf{T^{(i)}_{j};Y}$ is zero~\cite[Appendix~F.6]{Amjad_Understanding_arXiv}.\\

Information-theoretic quantities derived from \emph{individual} neuron outputs are not sufficient to characterize the behavior of a NN. Consider the following four examples:

\begin{example}\label{ex:entropy}
 Suppose that the NN has ReLU activation functions and that the quantizer $Q$ has two quantization bins, separated by a threshold at zero. Thus, $T_j^{(i)}$ indicates whether the neuron is active or not. Neurons that are active for half of the input samples have highest entropy. Ablating neurons with zero entropy $\ent{T_j^{(i)}}$ thus ablates neurons that are either inactive (``dead'') or active for all inputs, irrespective of the fact that the latter may play an important role in classification. Note further that neurons with small but non-zero $\ent{T_j^{(i)}}$ may also be important for classification if they are active for only one of many possible classes.
\end{example}

\begin{example}\label{ex:klred}
 Suppose that $C>2$ and that class $y\in\setnotation{C}$ is particularly easy to predict from the neuron outputs in the $i$-th layer. Suppose further that we use KL selectivity to measure neuron importance. It may happen that $y$ is the maximizer of~\eqref{eq:KLDclassmax} for all neurons in the layer. Thus, neuron importance is evaluated only based on the ability to distinguish class $y$ from the rest, which ignores separating the remaining classes. Ablating neurons based on KL selectivity may thus result in a NN unable to correctly classify classes other than $y$.
\end{example}

\begin{example}\label{ex:mired}
	Suppose the $j$-th and $k$-th neuron in the  $i$-th layer have the same output for every input $x$, i.e., $t_j^{(i)} = t_k^{(i)}$, and that $\mutinf{T_j^{(i)};Y}$ is large. If we use mutual information for ablation, then both neurons will be given high importance. Still, the neurons are redundant since one can always replace the other by adjusting the outgoing weights accordingly.
\end{example}

\begin{example}\label{ex:misynergy}
 Suppose that $C=2$ and that we use mutual information to measure neuron importance. Suppose further that the $j$-th and $k$-th neuron in the $i$-th layer are binary. It may happen that both $T_j^{(i)}$ and $T_k^{(i)}$ are independent of $Y$, but that $Y$ equals the \emph{exclusive or} of $T_j^{(i)}$ and $T_k^{(i)}$. Thus, $\mutinf{T_j^{(i)};Y}=\mutinf{T_k^{(i)};Y}=0$, although both neuron jointly determine the class, i.e., $\mutinf{T_j^{(i)}T_k^{(i)};Y}=\ent{Y}$.
\end{example}

In the first example, coarse quantization makes entropy an inadequate measure for the importance of a neuron for the classification task. In the second example, the neurons are individually informative, but KL selectivity may declare a set of neurons as important that is redundant (in the sense of determining class $y$) but insufficient (for determining other classes). Redundancy is also the phenomenon underlying the third example. In the fourth example, the neurons are individually uninformative, but jointly so -- the \emph{synergistic} information of this pair of neurons is high. While parts of these issues can be accounted for by finer quantization, PID, or the introduction of additional information-theoretic quantities (see Section~\ref{sec:discussion}), we show in the next section that cumulative neuron ablation based on even simple information quantities allows us to partially counteract some of the shortcomings mentioned in these examples. Namely, the effects in Examples~\ref{ex:entropy} through~\ref{ex:misynergy} become apparent in the classification results for information-ordered ablation, which allows us to formulate hypotheses about the activity, redundancy, and synergy of an entire layer.

\section{Understanding Neural Networks and Individual Neuron Importance via Cumulative Ablation}\label{sec:understanding}
To connect the proposed information-theoretic measures to the classification performance of trained NNs, we performed cumulative information-ordered ablation: Using the computed measures, we rank the neurons of each layer or of the NN as a whole. We subsequently cumulatively ablate the lowest- or the highest-ranking neurons and compute the classification error on the test dataset.  If cumulatively ablating neurons with low (high) values leads to small (large) drops in classification performance, then the information-theoretic measure on which the ranking is based can be assumed to indicate that the ablated neurons are important for good classification performance. In addition to that, the shape of the error curves thus obtained allows us to draw conclusions regarding the behavior of an entire layer, or of the network as a whole.

We chose cumulative ablation over the ablation of single neurons because most NNs used in practice are highly overparameterized and hence often exhibit high levels of redundancy. Ablating single neurons therefore has often only negligible effect on classification performance and hence fails to yield insights about the relation between the importance of the neuron for classification and its properties such as class selectivity. Cumulative ablation often has a greater impact on classification performance than the summed impact of single neuron ablation, as was also observed in~\cite{Meyes_Ablation} for pairs of neurons. 

We performed experiments with five different fully-connected NNs trained on three different datasets:
\begin{itemize}
 \item A $784-100-100-10$ NN trained on MNIST,
 \item a $784-100-100-10$ NN trained on FashionMNIST, 
 \item a $784-20-20-10$ NN trained on FashionMNIST (dubbed ``small FashionMNIST''),
 \item a $3072-250-500-250-500-10$ NN trained on CIFAR-10, and 
 \item a $3072-1000-750-500-250-10$ NN trained on CIFAR-10 (dubbed ``Funnel CIFAR-10'').
\end{itemize}

For each network architecture, we performed experiments with both ReLU and sigmoid activation functions. The NNs were trained in order to minimize cross-entropy without regularization, with $L_2$-regularization (weight decay $10^{-4}$), with Dropout in hidden layers (Dropout probabilities: $[0.3, 0.4]$ for MNIST and FashionMNIST, $[0.2, 0.4, 0.2, 0.4]$  for CIFAR-10), and with Dropout and batch normalization in the hidden layers. NNs were trained on 80\% of the designated training set, the remaining 20\% of the designated training set was used as validation set $\setnotation{D}$. Training was performed using the RMSProp optimizer with a learning rate of $0.001$, momentum of $0.01$ ($0.1$ for CIFAR-10) and a batch size of $32$. The number of training epochs was chosen individually for each dataset and regularizer setting, ranging between $8$ and $30$ epochs, to achieve good classification performance while avoiding overfitting.

Note that our goal for training was not to achieve state-of-the-art performance for fully-connected NNs for the considered datasets. Rather, our aim was to avoid factors such as overfitting, data augmentation, or linear bottleneck layers, that may confound our findings, while still achieving decent classification performance indicating that training was successful. During initial exploratory experiments, we observed that our qualitative results continue to hold for NNs for which the classification performance varies within a reasonable range. Moreover, the effects discussed below appear to be independent of the choice of optimizer and the optimizer parameters as long as the training converges without overfitting.

We computed the information-theoretic measures~\eqref{eq:ent},~\eqref{eq:MI}, and~\eqref{eq:KLDclassmax} for each neuron in each trained NN using the validation sets $\setnotation{D}$. Designing the quantizers for estimating information-theoretic quantities is challenging in general (cf.~recent discussions in~\cite{Saxe_IBTheory,Tishby_BlackBox}) but appears to be unproblematic in our case. We observed in our experiments that using more than two quantization bins did not yield significantly different results; we therefore discuss results for one-bit quantization, i.e., $|\setnotation{T}|=2$. Specifically, we selected the quantizer thresholds to lie at 0.5 and 0 for sigmoid and ReLU activation functions, respectively. Note that, with these settings and for ReLU activation functions, the activation value $0$ is identified with one quantization bin, while every positive activation is assigned to the second quantization bin. The variable $T_j^{(i)}$ thus indicates whether the output of a ReLU neuron is active or not (cf.~Example~\ref{ex:entropy}).

\begin{figure*}[t]
\subfigure[MNIST, ReLU, Dropout\label{fig:distribution:MNIST}]{
    {
    \includegraphics[width=0.45\textwidth,trim=3.5cm 0.2cm 3.5cm 1.2cm,clip]{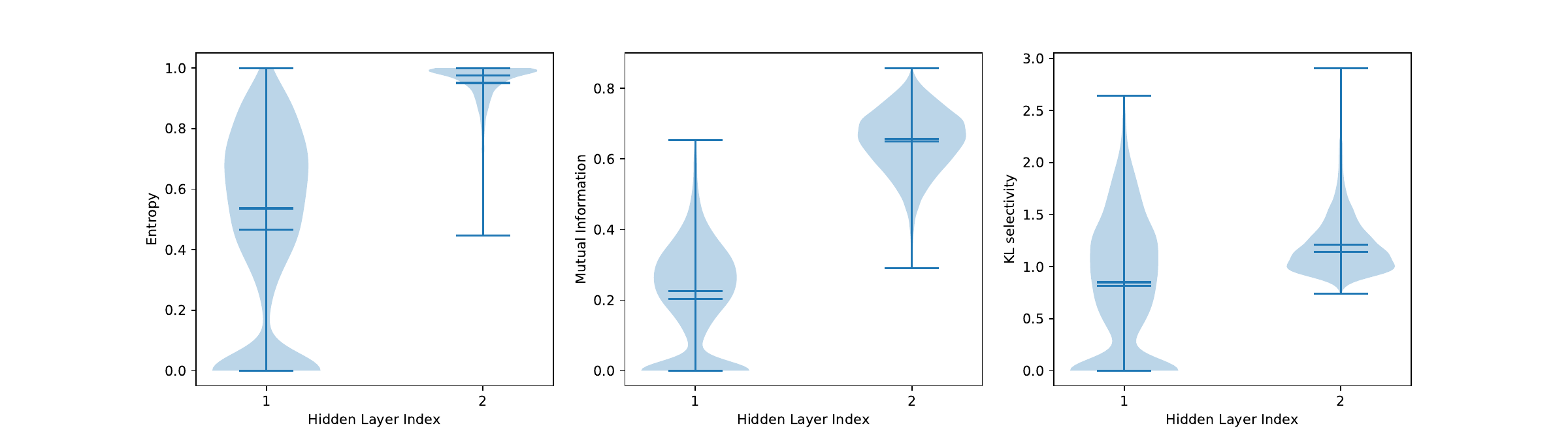}
    }
}\hfill
\subfigure[FashionMNIST, ReLU, $L_2$ regularization\label{fig:distribution:FashionMNIST}]{
    {
    \includegraphics[width=0.45\textwidth,trim=3.5cm 0.2cm 3.5cm 1.2cm,clip]{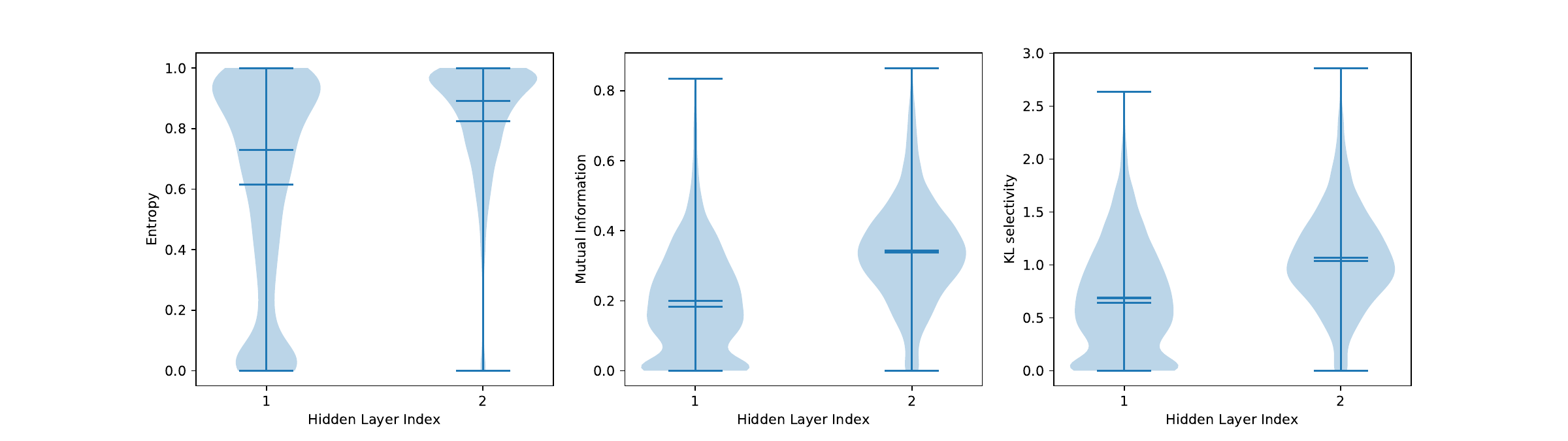}
    }
}
\subfigure[small FashionMNIST, ReLU, Dropout\label{fig:distribution:smallFashionMNIST}]{
    {
    \includegraphics[width=.45\textwidth,trim=3.5cm 0.2cm 3.5cm 1.2cm,clip]{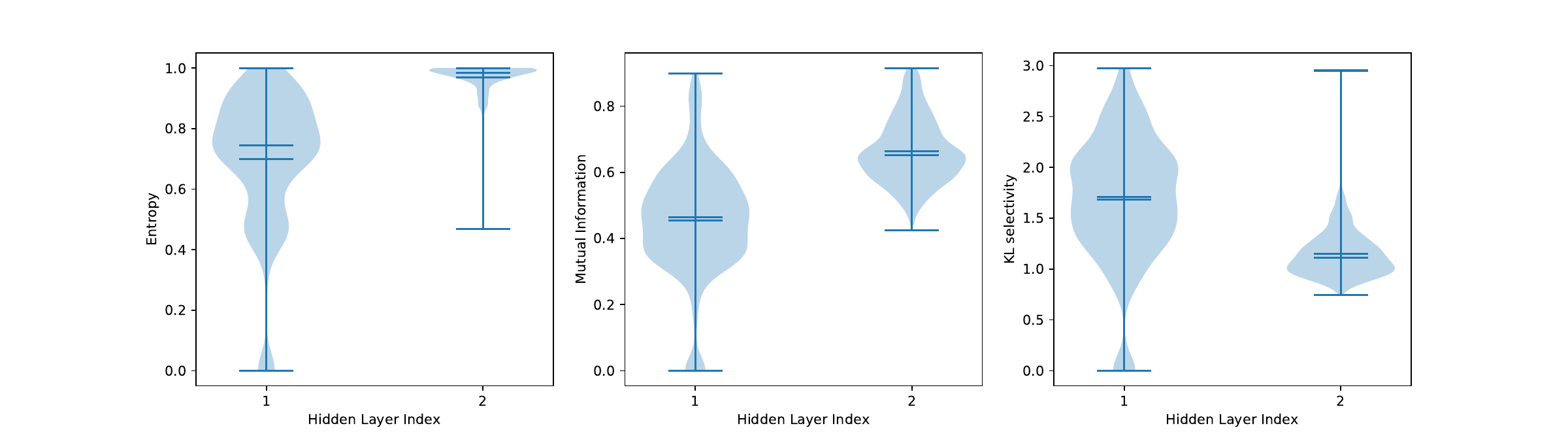}
    }
}\hfill
\subfigure[CIFAR10, ReLU, $L_2$ regularization\label{fig:distribution:CIFAR}]{
    {
    \includegraphics[width=0.45\textwidth,trim=3.5cm 0.2cm 3.5cm 1.2cm,clip]{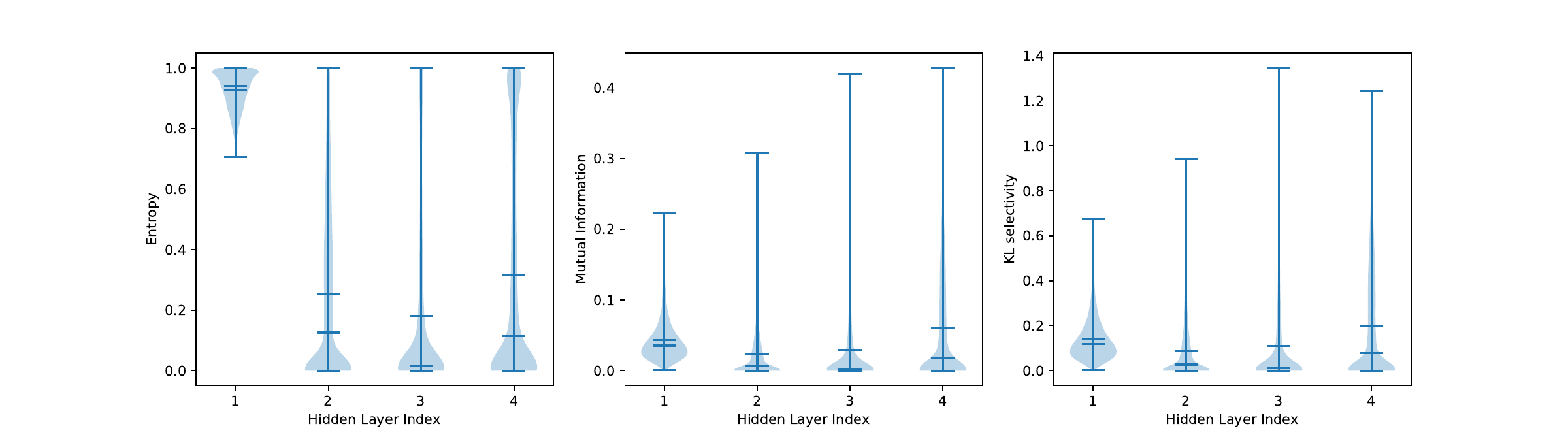}
    }
}
\caption{Distribution of information-theoretic measures for neurons in different layers, visualized as violin plots. The vertical lines indicate minimum and maximum values, as well as the means and the medians of the distributions. Neurons in deeper layer have larger entropy, mutual information, and KL selectivity. These results are generated by combining the results of all randomly initialized NNs.}
\label{fig:distribution}
\end{figure*}

Following~\cite{Deepmind_Cats}, we perform ablation to zero, i.e., we replace the output of each ablated neuron by zero, in order to make our results and conclusions comparable. In a few cases, we observed that ablation to the mean, i.e., replacing the output of each ablated neuron by its mean value on the training set, led to slightly increased robustness against cumulative ablation.

All of the results shown in this paper are obtained by training $50$ NNs using the same settings but with different random initializations for MNIST and FashionMNIST, and $35$ NNs for CIFAR-10, respectively. The ablation curves show the means and standard deviations of these experiments. To compare different ablation curves, we performed statistical tests. More specifically, when comparing information-ordered with random ablation, we report the mean gap (MG) at a given number of ablated neurons together with its $p$-value obtained using a Wilcoxon signed-rank test. When comparing random ablation curves for two different experimental scenarios, we follow the same procedure, but use a two-sample Kolmogorov-Smirnov test instead. For example, for a mean gap of 5 percentage points after pruning 80 neurons resulting in a $p$-value of 0.1, we write \stat{80}{5}{0.1}. We call the behavior of ablation curves similar if either the mean gap is small or if $p>0.05$. The models are implemented\footnote{The source code of our experiments can be downloaded from \url{https://raa2463@bitbucket.org/raa2463/neuron-importance-via-information.git}.} in Python using Pytorch.

\subsection{Neurons Become more Variable, Informative, and Selective Towards Deeper Layers}\label{subsec:layerdistribution}

Fig.~\ref{fig:distribution} shows the empirical distribution of the proposed information-theoretic quantities for different layers of the trained NNs. It can be observed that, in general, all quantities increase towards deeper layers. This is in agreement with observations in~\cite[Figs.~A2~\&~A4.a]{Deepmind_Cats},~\cite[Fig.~1E]{Bau_Understanding}, and~\cite{Yu_CriticalPaths} suggesting that shallow layers are \emph{general}, i.e., not related to a specific class, whereas features in deeper layers are more and more specific towards the class variable. For example, that the mutual information terms $\mutinf{T_j^{(i)};Y}$, corresponding to the individual neuron's quantized outputs, increase towards deeper layers suggests that neurons become informative about the class variable individually rather than collectively; i.e., while in the shallow layers class information can only be obtained from the joint observation of multiple neurons encoding general features, in deeper layers individual neurons may represent features that are informative about a set of classes.

We have confirmed the same trend for all information-theoretic quantities and all combinations of activation functions and regularization techniques for MNIST, and for entropy and mutual information for FashionMNIST, respectively.
The behavior for the NN trained on CIFAR-10 is less obvious, cf.~Fig~\ref{fig:distribution:CIFAR}; however, one still observes a small increase of the relevant quantities from the second to the fourth hidden layer. We believe that the behavior for CIFAR-10 can be explained by many neurons in these layers being inactive or uninformative, respectively. We will touch upon this issue again in Section~\ref{subsec:layerpruning:CIFAR}. Finally, while small FashionMNIST with $L_2$ regularization shows similar qualitative behavior as Fig.~\ref{fig:distribution:FashionMNIST}, training with Dropout leads to the second hidden layer having smaller KL selectivity than the first. Since, in contrast, with finer quantization the KL selectivity values of the first and second hidden layer seem to have similar distribution, we believe that this is an artifact of our coarse quantization strategy. With both fine and coarse quantization, we have observed that mutual information values are larger for training with Dropout than with $L_2$ regularization. This indicates that Dropout encourages the information to spread more evenly over the few available neurons, thus making them redundant.
 
The behavior of mutual information in Fig.~\ref{fig:distribution} appears to be in contrast with the behavior of the mutual information between the class and the complete layer, i.e., with $\mutinf{\{t_j^{(i)}\};Y}$. The data processing inequality (cf.~\cite[Th.~2.8.1]{Cover_Information}) dictates that $\mutinf{\{t_j^{(i)}\};Y}$ should decrease towards deeper layers; proper training reduces this decrease, as empirically observed in~\cite{Tishby_BlackBox,Saxe_IBTheory}. That the terms $\mutinf{T_j^{(i)};Y}$ increase towards deeper layers suggests that neurons in deeper layers exhibit a higher degree of redundancy. We will touch up this in Sections~\ref{subsec:layerpruning:MNIST} and~\ref{sec:discussion}.

\subsection{Whole-Network Cumulative Ablation Analysis}\label{subsec:pruning}
\begin{figure}[t]
\begin{center}
    \subfigure[MNIST, $L_2$ regularization \label{fig:wholeNNpruning:MNISTl2}]
    {
        \includegraphics[width=0.23\textwidth,trim=0.5cm 0 1.2cm 0,clip]
        {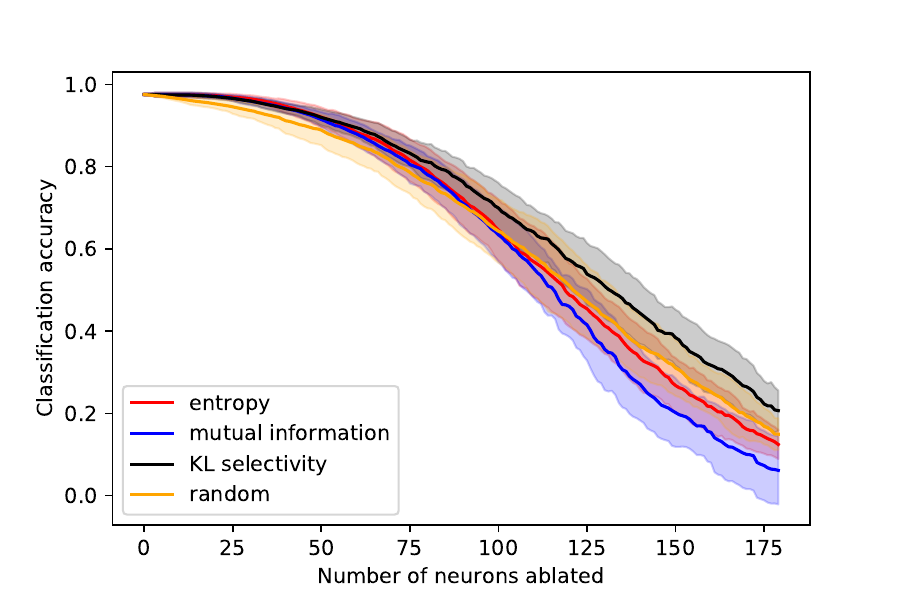}
    }\hfill
    \subfigure[FashionMNIST, $L_2$ regularization \label{fig:wholeNNpruning:FashionMNISTl2}]
    {
        \includegraphics[width=0.23\textwidth,trim=0.5cm 0 1.2cm 0,clip]
        {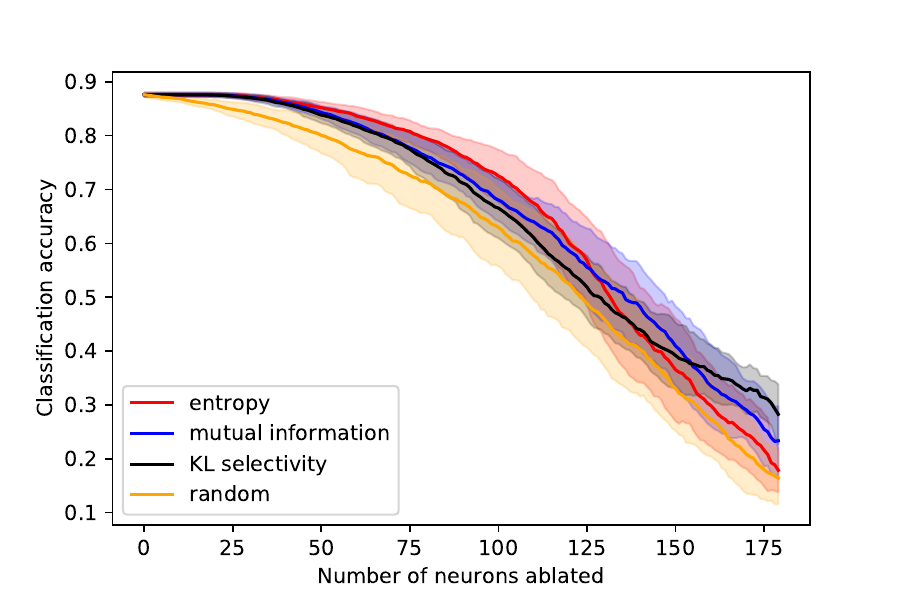}
    }
    \subfigure[MNIST, Dropout \label{fig:wholeNNpruning:MNISTdropout}]
    {
        \includegraphics[width=0.23\textwidth,trim=0.5cm 0 1.2cm 0,clip]
        {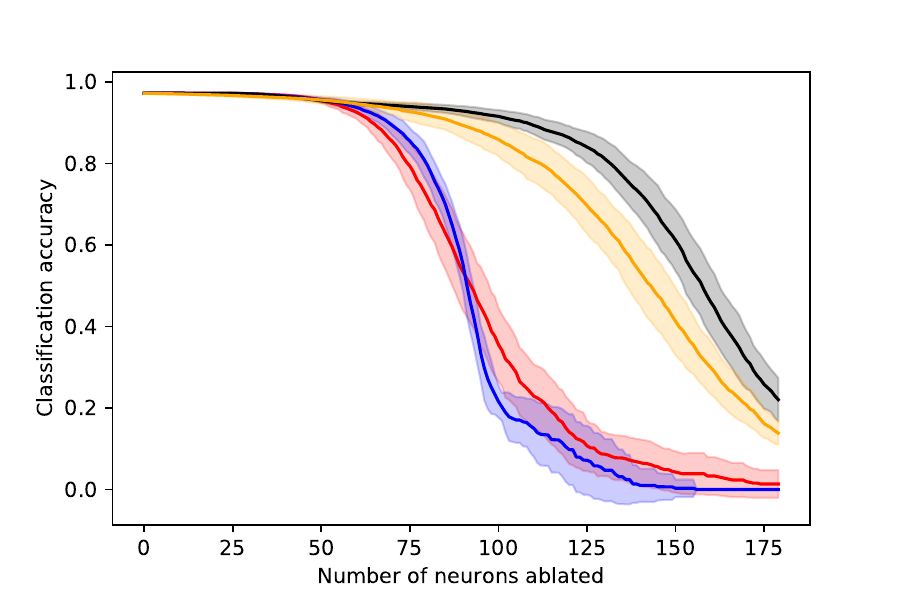}
    }\hfill
    \subfigure[FashionMNIST, Dropout \label{fig:wholeNNpruning:FashionMNISTdropout}]
    {
        \includegraphics[width=0.23\textwidth,trim=0.5cm 0 1.2cm 0,clip]
        {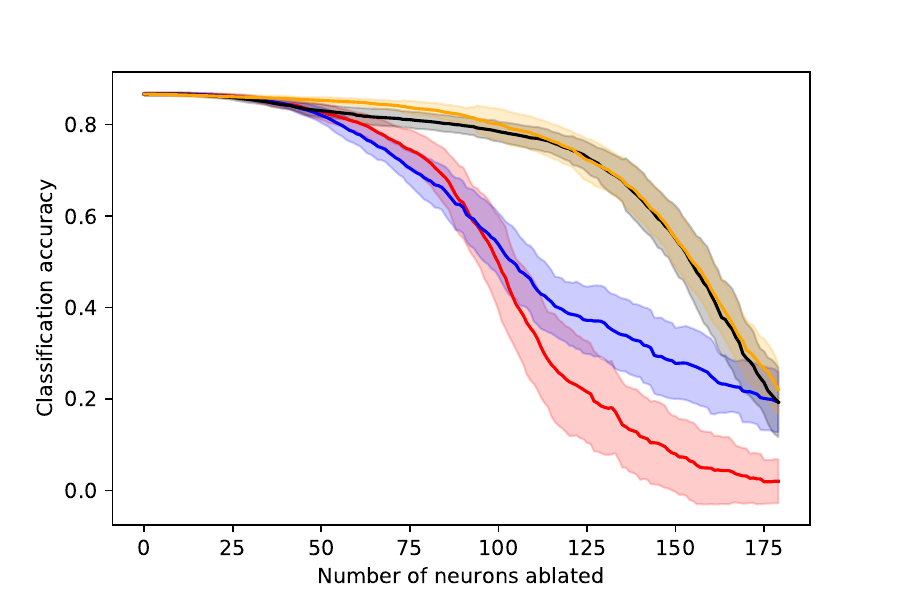}
    }
    \subfigure[MNIST, $L_2$ regularization, ablation to the mean \label{fig:wholeNNpruning:MNISTsigmoid}]
    {
        \includegraphics[width=0.23\textwidth,trim=0.5cm 0 1.2cm 0,clip]
        {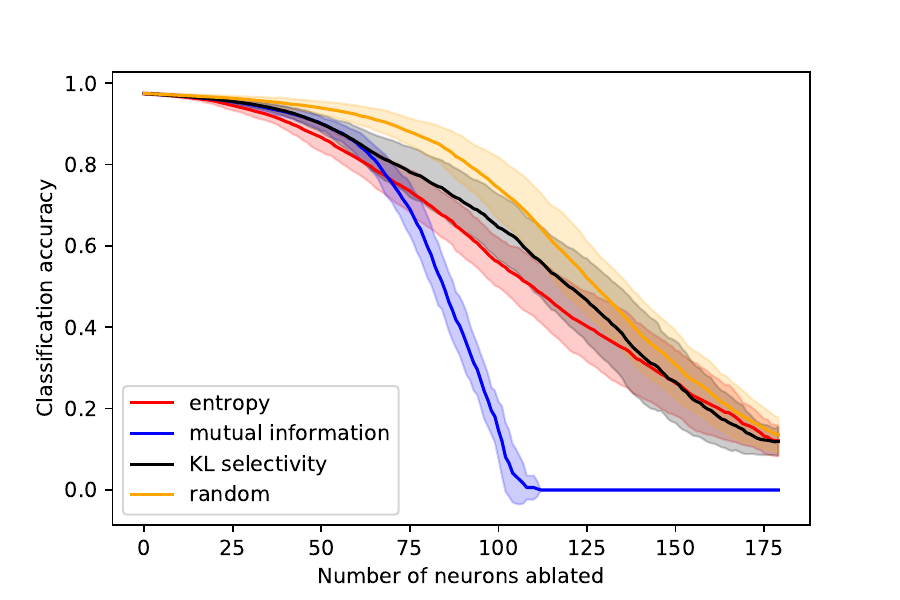}
    }\hfill
    \subfigure[FashionMNIST \label{fig:wholeNNpruning:FashionMNISTsigmoid}]
    {
        \includegraphics[width=0.23\textwidth,trim=0.5cm 0 1.2cm 0,clip]
        {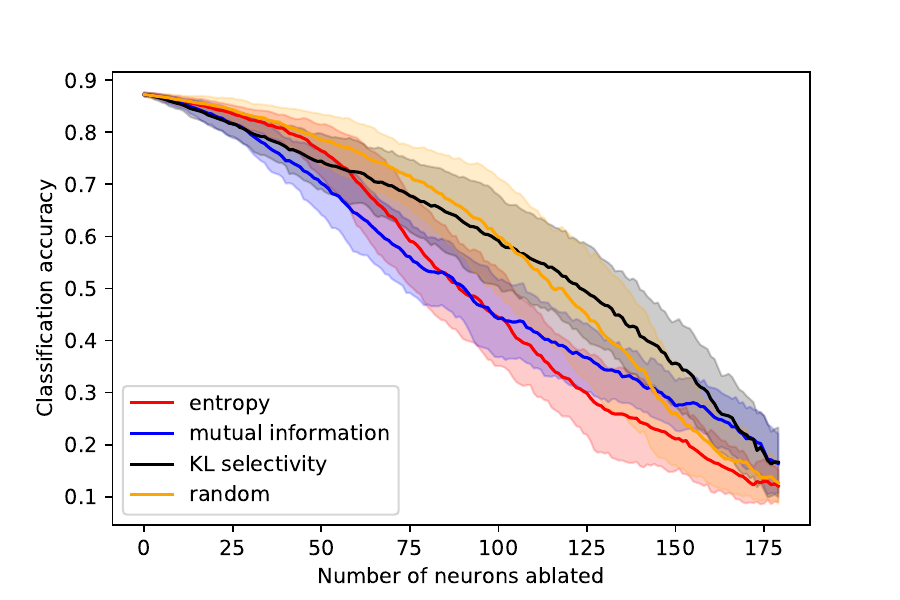}
    }
    \caption{Cumulative ablation (neurons with low information-theoretic quantities are ablated first) in NNs with ReLU (a-d) and sigmoid (e-f) activation functions. Ablation across all layers indicates that information-theoretic quantities have little correlation (a,c) or negative correlation (b,d) with classification performance. The lines and shaded regions represent the mean and standard deviation over all randomly initialized NNs, respectively.}
    \label{fig:wholeNNpruning}
\end{center}
\end{figure}

The authors of~\cite{Deepmind_Cats} concluded that neither mutual information nor class selectivity are correlated with classification performance, and that highly selective neurons may actually harm classification. To review this claim, we rank all neurons in the NNs with ReLU activation functions trained on MNIST and FashionMNIST w.r.t.\ their information-theoretic quantities and ablate to zero those neurons with lowest ranks (i.e., we perform cumulative ablation analysis across both layers simultaneously). The results are shown in Fig.~\ref{fig:wholeNNpruning}. For NNs with ReLU activation functions trained with $L_2$ regularization, ablating neurons with small entropy or mutual information values performs similarly as ablating neurons randomly (Figs.~\ref{fig:wholeNNpruning:MNISTl2} and~\ref{fig:wholeNNpruning:FashionMNISTl2}). In contrast, in NNs trained with Dropout or in NNs with sigmoid activation functions, ablating neurons with small entropy or mutual information values performs worse (Figs.~\ref{fig:wholeNNpruning:MNISTdropout} to~\ref{fig:wholeNNpruning:FashionMNISTsigmoid}). For example, when ablating neurons with small entropy values rather than randomly, we obtain \stat{101}{-0.19}{0.93} in Fig.~\ref{fig:wholeNNpruning:MNISTl2} but \stat{101}{-51.2}{7.55 \cdot 10^{-10}} in Fig.~\ref{fig:wholeNNpruning:MNISTdropout}. This indicates that entropy and mutual information are at best weakly positive or even negatively correlated with classification performance, respectively.
The latter effect was also observed in~\cite[Fig.~A4.a]{Deepmind_Cats}, where the authors claimed that neurons with large mutual information have adverse effects on classification performance. Judging from Fig.~\ref{fig:wholeNNpruning}, also KL selectivity seems to be largely uncorrelated with classification performance: ablating neurons with low KL selectivity performs similarly as ablating neurons randomly (\stat{125}{0.3}{0.78} in Fig.~\ref{fig:wholeNNpruning:FashionMNISTdropout}).
The results are thus in line with those of~\cite{Deepmind_Cats} and suggest that it is not the neurons with largest KL selectivity or mutual information values within the entire NN that are important for classification.

We now provide a new interpretation in the light of the results of Section~\ref{subsec:layerdistribution}.
With reference to Figs.~\ref{fig:distribution:MNIST} and~\ref{fig:distribution:FashionMNIST}, ablating neurons with lowest mutual information mostly ablates neurons in the first hidden layer. These neurons extract general features that are combined in the second layer to features that are specific to the class variable, leading to higher mutual information and KL selectivity in the second layer. By ablation, these generic features are removed, thus deeper layers are not able to extract class-specific features anymore and classification suffers. The effect is most strongly pronounced for Dropout, where most neurons in the first layer have smaller mutual information than any neuron in the second layer (see Fig.~\ref{fig:distribution:MNIST}). Neurons are thus ablated almost exclusively from the first layer, which explains why classification performance drops so quickly in Fig.~\ref{fig:wholeNNpruning:MNISTdropout} when the number of neurons ablated approaches the size of the first hidden layer. The same effect holds also for entropy and KL selectivity, and more generally, for most experiments with MNIST and FashionMNIST datasets. We thus conclude that, due to large differences between the information-theoretic quantities of different layers, cumulatively ablating neurons from multiple layers simultaneously cannot be used decide whether any of these proposed importance measures is a good indicator for classification performance.

\subsection{Layer-Wise Cumulative Ablation Analysis: MNIST}
\label{subsec:layerpruning:MNIST}

\begin{figure}[t]
\begin{center}
    \subfigure[ReLU, Dropout, Layer 1 (left) and Layer 2 (right) \label{fig:layerpruningMNIST:Relu}]
    {
        \includegraphics[width=0.23\textwidth,trim=0.5cm 0 1.2cm 0,clip]
        {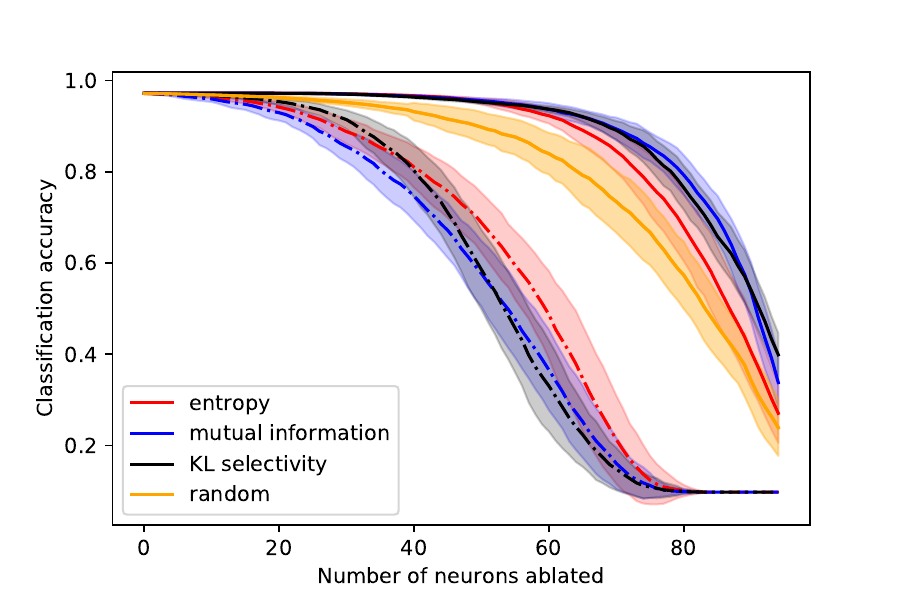}
        \includegraphics[width=0.23\textwidth,trim=0.5cm 0 1.2cm 0,clip]
        {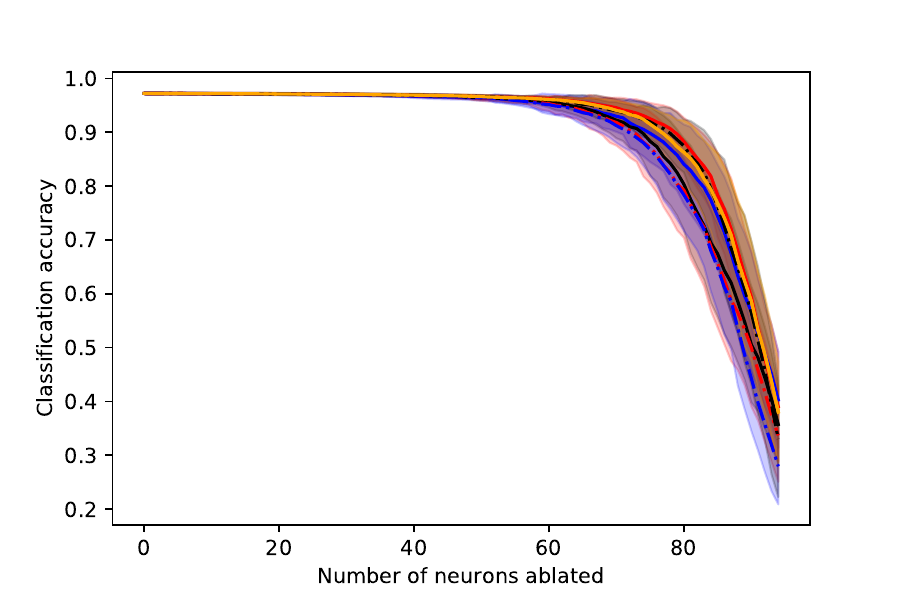}
    }   
    \subfigure[ReLU, Batchnorm and Dropout, Layer 1 (left) and Layer 2 (right)\label{fig:layerpruningMNIST:batchnorm}]
    {
        \includegraphics[width=0.23\textwidth,trim=0.5cm 0 1.2cm 0.5cm,clip]
        {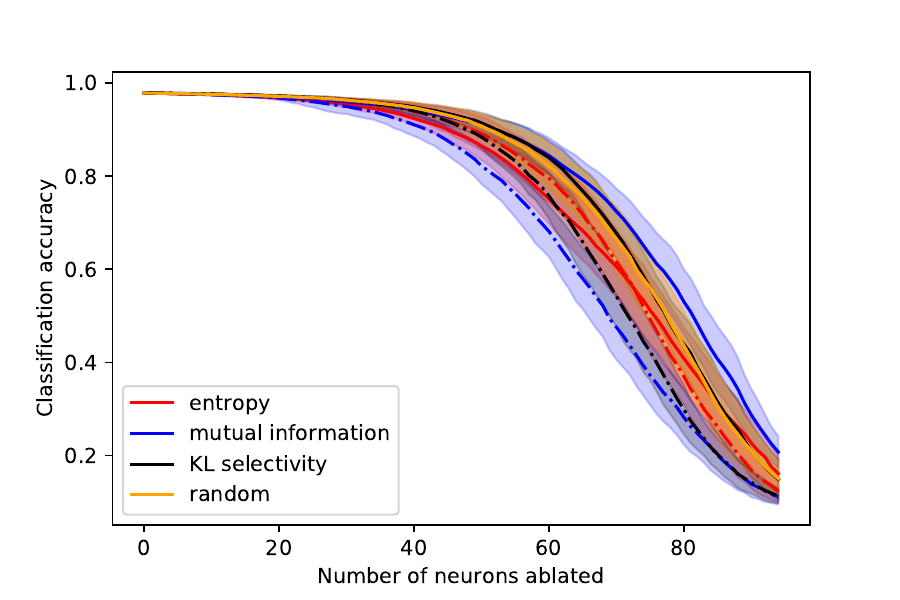}
        \includegraphics[width=0.23\textwidth,trim=0.5cm 0 1.2cm 0.5cm,clip]
        {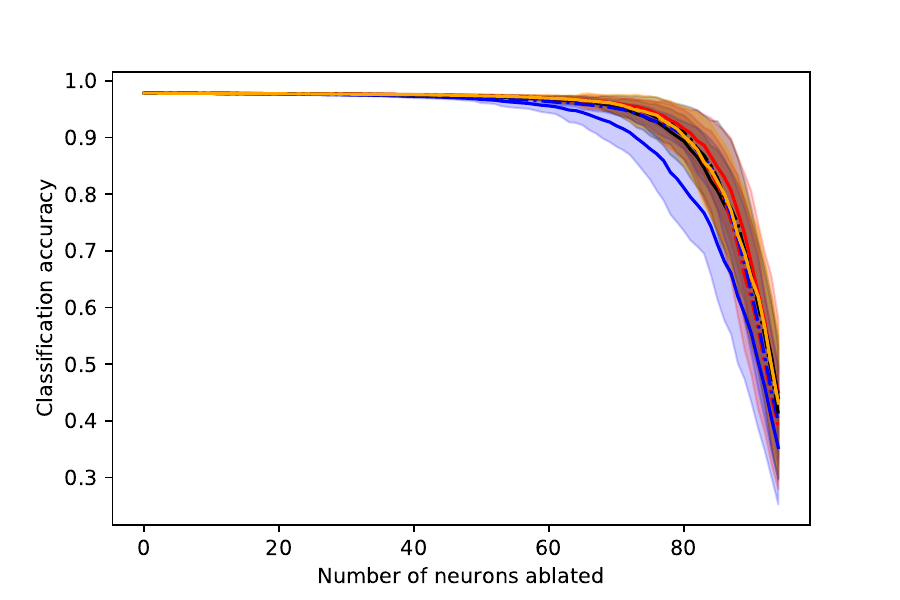}
    }
    \subfigure[Sigmoid, $L_2$ regularization, Layer 1 (left) and Layer 2 (right)\label{fig:layerpruningMNIST:sigmoid}]
    {
        \includegraphics[width=0.23\textwidth,trim=0.5cm 0 1.2cm 0,clip]
        {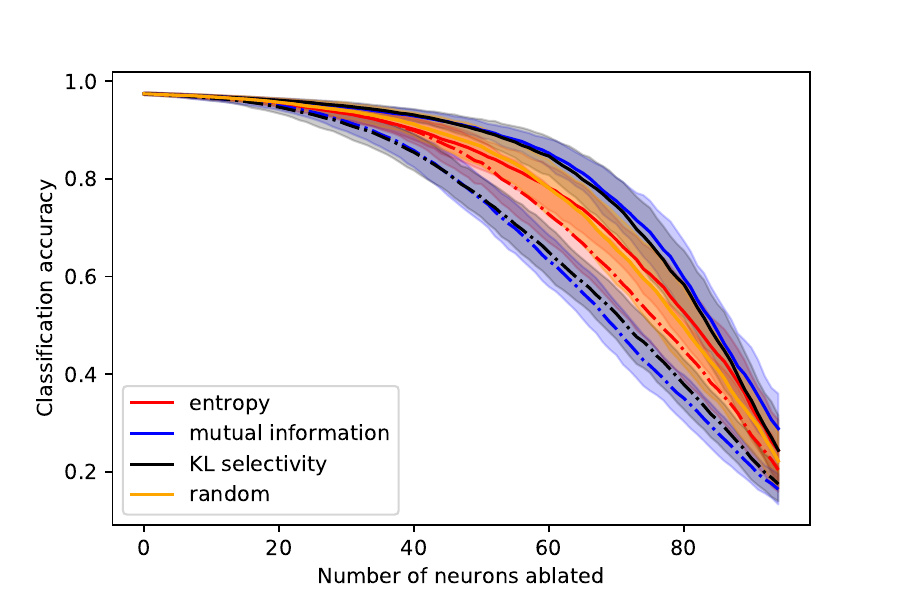}
        \includegraphics[width=0.23\textwidth,trim=0.5cm 0 1.2cm 0,clip]
        {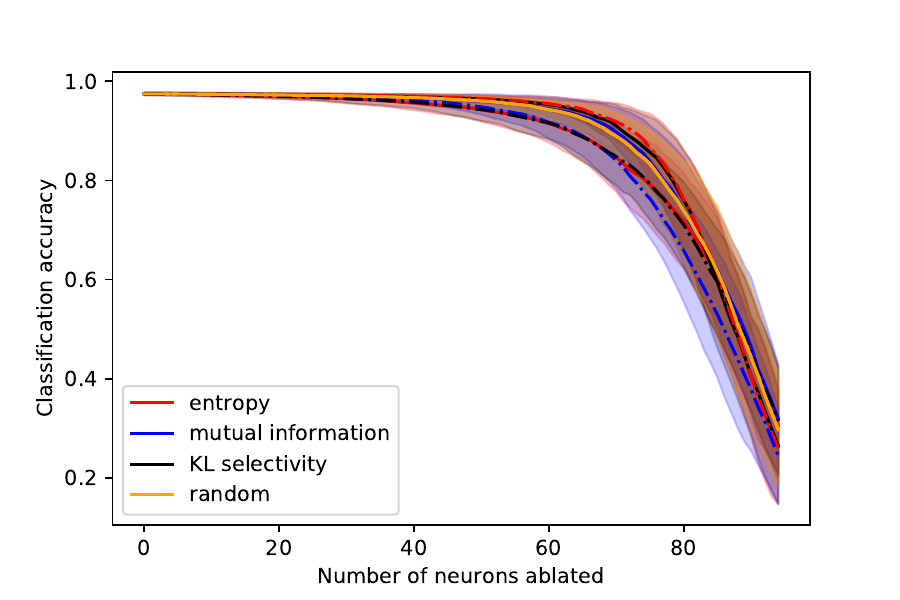}
    }
    \subfigure[Sigmoid, Dropout, Layer 1 (left) and Layer 2 (right)\label{fig:layerpruningMNIST:sigmoid:dropout}]
    {
        \includegraphics[width=0.23\textwidth,trim=0.5cm 0 1.2cm 0,clip]
        {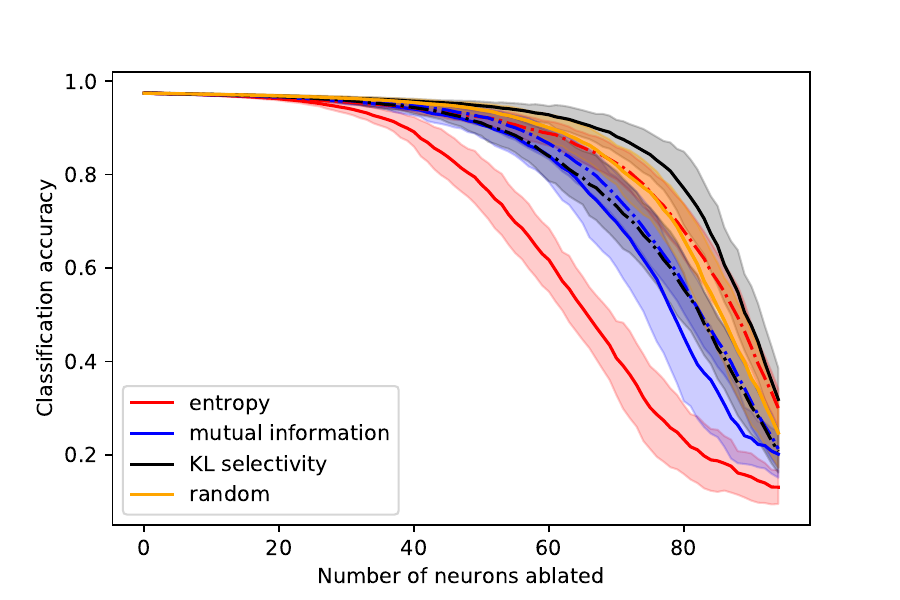}
        \includegraphics[width=0.23\textwidth,trim=0.5cm 0 1.2cm 0,clip]
        {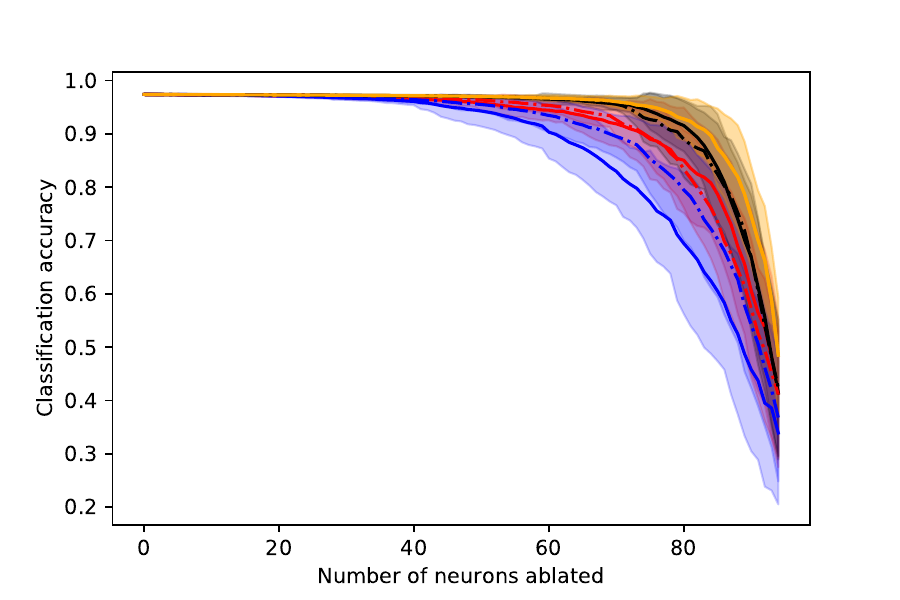}
    }
    \caption{Cumulative ablation of neurons in the first (left plots) and second (right plots) layer of a NN trained on MNIST. Neurons with low (solid) or high (dashed) information-theoretic quantities are ablated. The neurons in (c) were ablated to the mean.}
    \label{fig:layerpruningMNIST}
\end{center}
\vspace{-.5cm}
\end{figure}

Since the conclusion that neurons with large mutual information adversely affect classification performance appears counter-intuitive, we next perform ablation analysis in each layer separately. We start with presenting the results for a 784-100-100-10 network trained on the MNIST dataset.

The results are shown in Fig.~\ref{fig:layerpruningMNIST}. It can be seen that ablating neurons in the first hidden layer has stronger negative effects than ablating neurons in deeper layers (\stat{80}{-28.9}{1.98 \cdot 10^{-29}} for the random ablation curves in Fig.~\ref{fig:layerpruningMNIST:Relu}). For example, in Figs.~\ref{fig:layerpruningMNIST:Relu} and~\ref{fig:layerpruningMNIST:sigmoid}, ablating the 50 lowest-ranked neurons in the second hidden layer has negligible effect on classification performance. We believe that this is because many neurons in the second layer are redundant; the alternative that many neurons in the second layer are inactive or irrelevant for classification can be ruled out because of large entropy, mutual information, and KL selectivity values, cf.~Fig.~\ref{fig:distribution:MNIST}.

Most importantly, one can see in Fig.~\ref{fig:layerpruningMNIST} that ablating neurons with low (high) ranks from the first hidden layer leads to better (worse) classification performance than ablating neurons randomly, and the effect appears to be more pronounced in NNs with ReLU activation functions than in NNs with sigmoid activation functions. Ablating 80 neurons with low or high mutual information in Fig.~\ref{fig:layerpruningMNIST:Relu} yields a mean gap of 21.9\% ($p=7.55 \cdot 10^{-10}$) and -47.4\% ($p=7.55 \cdot 10^{-10}$), respectively, when compared to random ablation. These numbers reduce to \stat{80}{10}{1.6\cdot 10^{-5}} and \stat{80}{-11}{9\cdot 10^{-7}} in Fig.~\ref{fig:layerpruningMNIST:sigmoid}. This effect appears to be missing altogether if neurons are ablated from the second layer; we believe that this is again linked to the fact that the neurons in the second layer are highly redundant for this simple dataset. If, in NNs with sigmoid activation functions, neurons are ablated to zero, the results are inconclusive: The correlation between the candidate neuron importance measures and classification performance seems generally weak in these cases, and is moreover inconsistent over different regularizers. For example, for the NN trained with Dropout, ablating neurons with low entropy values from the first layer performs worse than ablating neurons randomly (\stat{80}{-42.8}{7.55\cdot 10^{10}}), and the effect of other ablation strategies appears to be small when compared to random ablation (see Fig.~\ref{fig:layerpruningMNIST:sigmoid:dropout}). Similar to results with sigmoid activation functions, our results with NNs trained using a combination of batch normalization and Dropout (referred to as ``batchnorm'' henceforth) are inconclusive, irrespective of the activation function. While the performance of these NNs on the classification task competes with the performance of NNs trained differently, the robustness against cumulative ablation seems not to be linked to the proposed importance measures, as information-ordered ablation yields similar results as random ablation. Indeed, ablating 80 neurons from the first hidden layer with high KL selectivity in Fig.~\ref{fig:layerpruningMNIST:batchnorm} reduces accuracy by only 13 percentage points ($p=9.63 \cdot 10^{-10}$), while ablating 80 neurons with low KL selectivity performs on par with random ablation (\stat{80}{0.3}{0.82}). We believe that the lack of such a correlation may again be due to our comparably coarse quantization procedure, combined with the fact that batchnorm has an immediate effect on activation values. Since we have observed similar results for different datasets and architectures, for the remainder of this paper we will consider NNs with ReLU activation functions and trained without batch normalization only. For further results for networks with sigmoid activation functions and/or trained with batch normalization, we refer the reader to~\cite[Appendices~C~\&~D]{Amjad_Understanding_arXiv}.

\subsection{Layer-Wise Cumulative Ablation Analysis: FashionMNIST}
\label{subsec:layerpruning:FashionMNIST}
\begin{figure}[t]
\begin{center}
    \subfigure[FashionMNIST, $L_2$ regularization, Layer 1 (left) and Layer 2 (right)\label{fig:layerpruningFashionMNIST:L2}]
    {
        \includegraphics[width=0.23\textwidth,trim=0.5cm 0 1.2cm 0,clip]
        {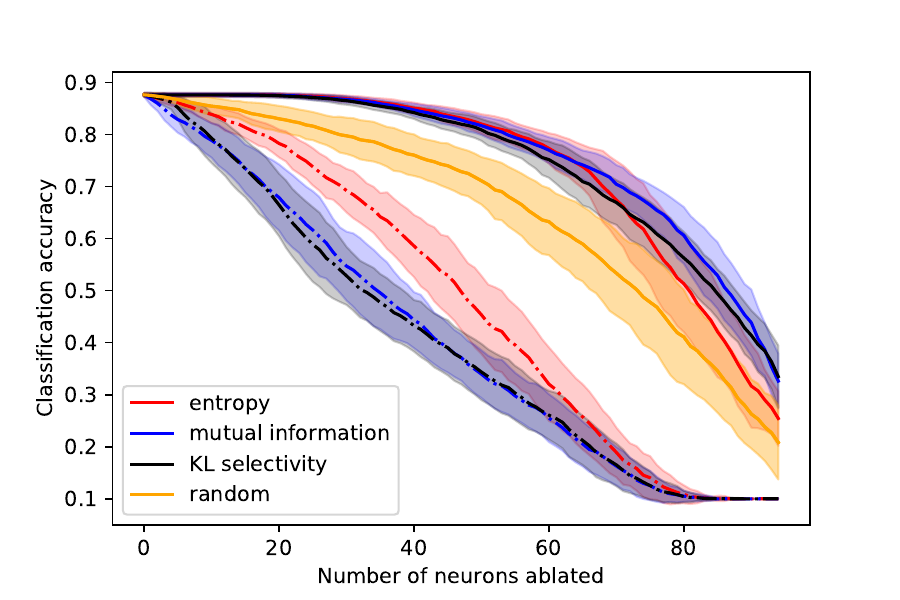}
        \includegraphics[width=0.23\textwidth,trim=0.5cm 0 1.2cm 0,clip]
        {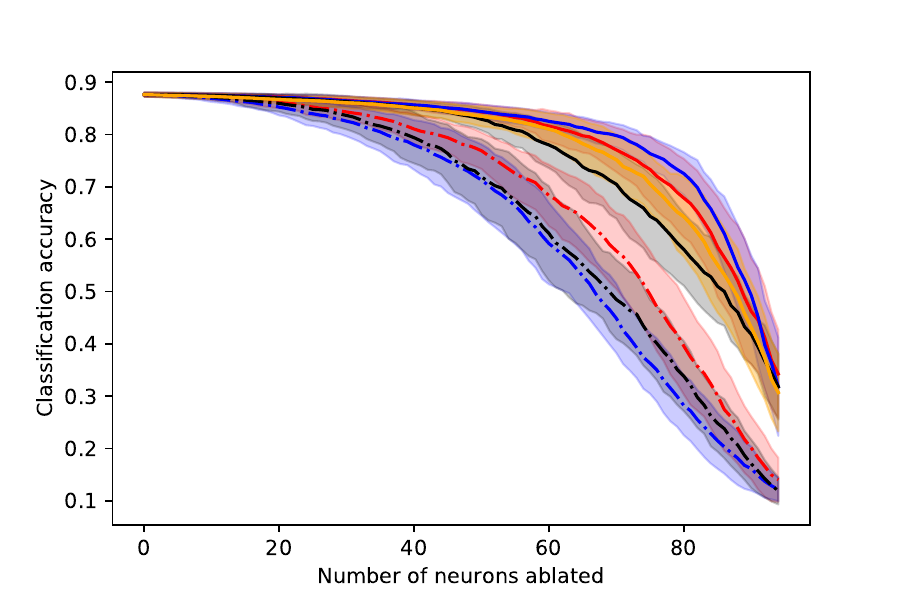}
    }
    \subfigure[Small FashionMNIST, $L_2$ regularization, Layer 1 (left) and Layer 2 (right) \label{fig:layerpruningFashionMNIST:small}]
    {
        \includegraphics[width=0.23\textwidth,trim=0.5cm 0 1.2cm 0,clip]
        {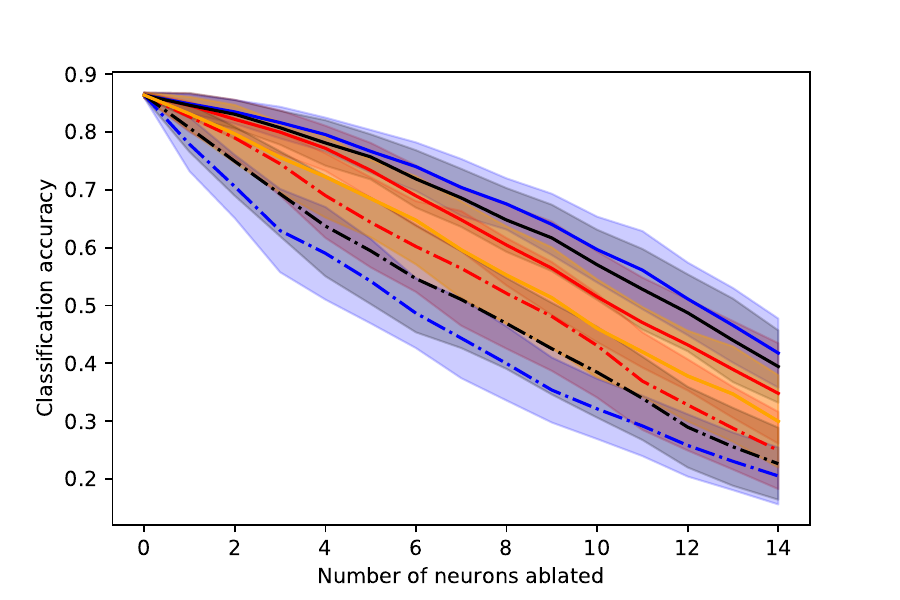}
        \includegraphics[width=0.23\textwidth,trim=0.5cm 0 1.2cm 0,clip]
        {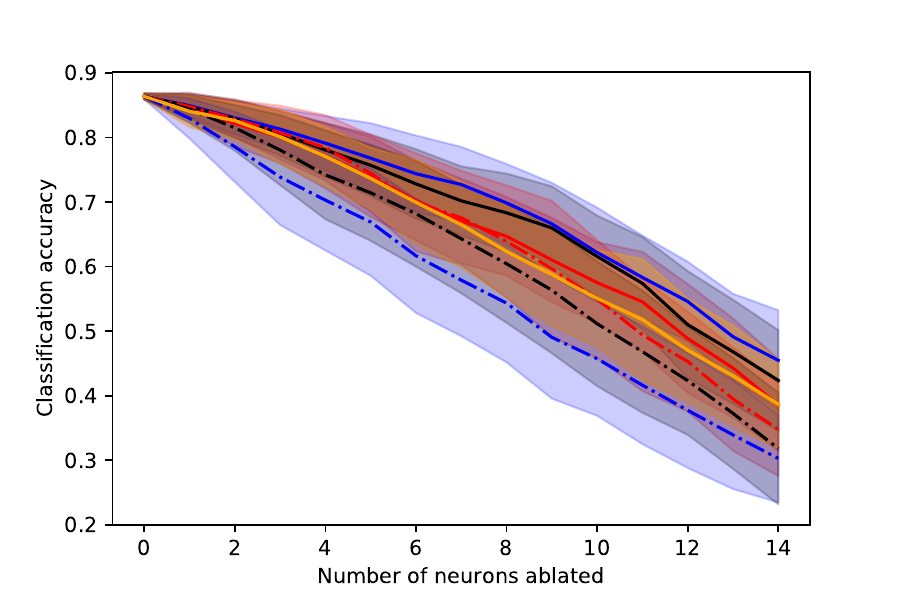}
    }
    \subfigure[Small FashionMNIST, Dropout, Layer 1 (left) and Layer 2 (right) \label{fig:layerpruningFashionMNIST:Dropsmall}]
    {
        \includegraphics[width=0.23\textwidth,trim=0.5cm 0 1.2cm 0,clip]
        {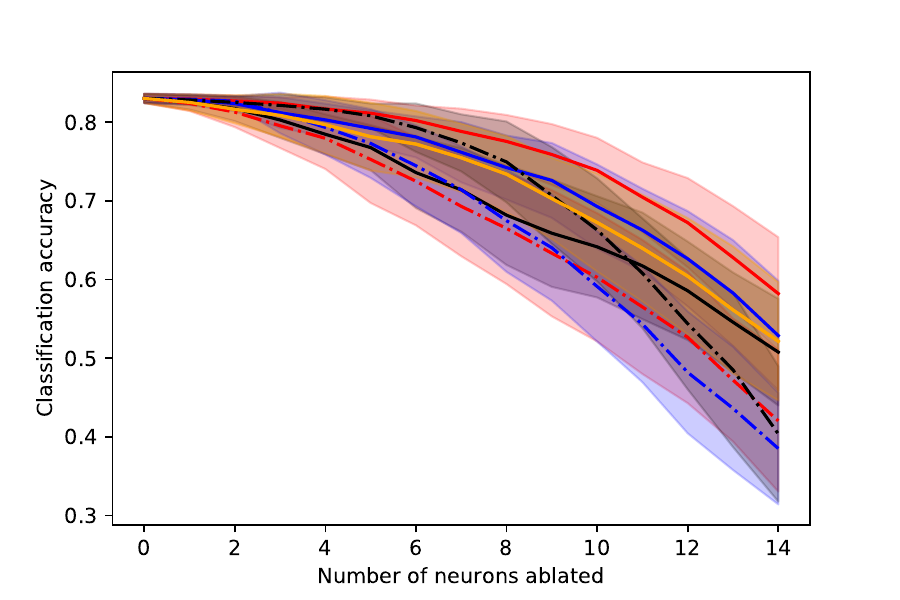}
        \includegraphics[width=0.23\textwidth,trim=0.5cm 0 1.2cm 0,clip]
        {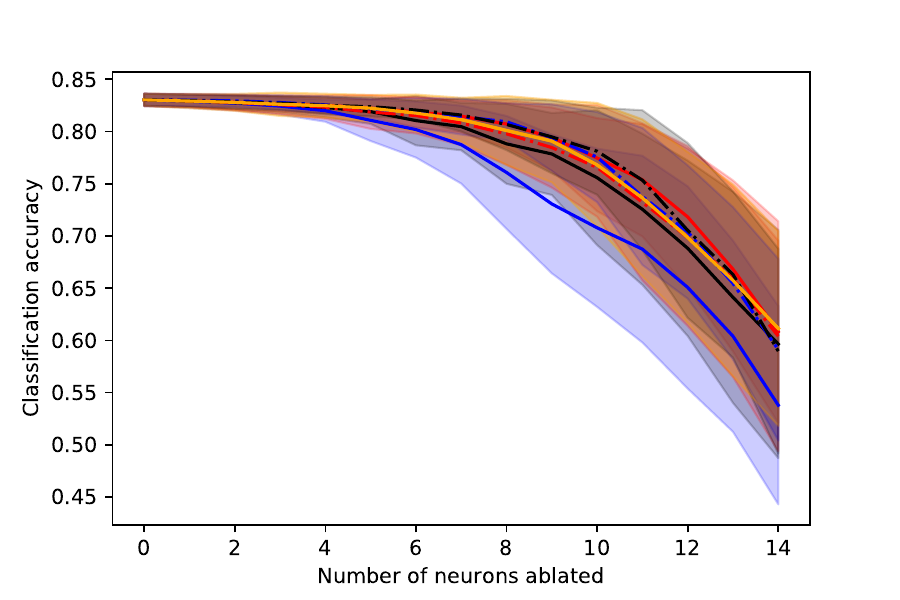}
    }
    \caption{Cumulative ablation of neurons in the first (left plots) and second (right plots) layer of a NN with ReLU activation functions trained on FashionMNIST.}
    \label{fig:layerpruningFashionMNIST}
\end{center}
\end{figure}

Fig.~\ref{fig:layerpruningFashionMNIST} summarizes our experiments with the FashionMNIST dataset. It can be seen that, again, the effects of cumulative ablation are more severe in the first hidden layer -- this effect is pronounced more strongly in the large architecture, since this architecture allows to create redundancy in the second layer (\stat{80}{-23.2}{3.68 \cdot 10^{-18}} in Fig.~\ref{fig:layerpruningFashionMNIST:L2}). Training the networks with $L_2$ regularization or without regularization leads to a positive correlation between the candidate importance measures and classification performance: Ablating neurons with large (small) mutual information, KL selectivity, or entropy affects classification accuracy more (less) negatively than ablating neurons randomly. Specifically, by ablating neurons from the first hidden layer with low or high KL selectivity values in Fig.~\ref{fig:layerpruningFashionMNIST:L2} yields \stat{80}{15.2}{3.56 \cdot 10^{-9}} and \stat{80}{-30.7}{7.55 \cdot 10^{-10}}, respectively, compared to random ablation. The effect is again only weak for entropy, which is independent of the class label variable, and strongest for mutual information. The correlation appears to be weaker (but still positive) for the small architecture, for which cumulative ablation has a stronger negative effect, as expected. In this small network architecture, hardly any redundancy is present, which makes almost all neurons important for classification.

Training the small architecture with Dropout increases robustness against cumulative ablation. However, as one can see in Fig.~\ref{fig:layerpruningFashionMNIST:Dropsmall}, ablating neurons with low (high) KL selectivity performs worse (better) than ablating neurons randomly. This suggests that class-specific neurons in the first layer actually harm classification performance: The restricted architecture is too small to benefit from class-specificity, but rather requires neurons in the first layer that contain information about multiple classes simultaneously. Neurons with the lowest mutual information seem to be important also in the second hidden layer; comparing Fig.~\ref{fig:distribution:smallFashionMNIST} with Fig.~\ref{fig:distribution:FashionMNIST}, one can see that these neurons contain more class information than most neurons in the second layer of the large architecture trained with $L_2$ regularization. Therefore, we believe that all neurons in the second layer are important for classification, and that this negative correlation between mutual information and classification performance is spurious. This conclusion is also supported by the fact that all other curves in Fig.~\ref{fig:layerpruningFashionMNIST:Dropsmall} almost overlap.

\subsection{Layer-Wise Cumulative Ablation Analysis: CIFAR-10}
\label{subsec:layerpruning:CIFAR}

\begin{figure*}[ht]
\begin{center}
    \subfigure[CIFAR-10, ReLU, $L_2$ regularization, Layers 1 (left) to 4 (right) \label{fig:layerpruningCIFAR:l2}]
    {
        \includegraphics[width=0.24\textwidth,trim=0.5cm 0 1.2cm 0.5cm,clip]
        {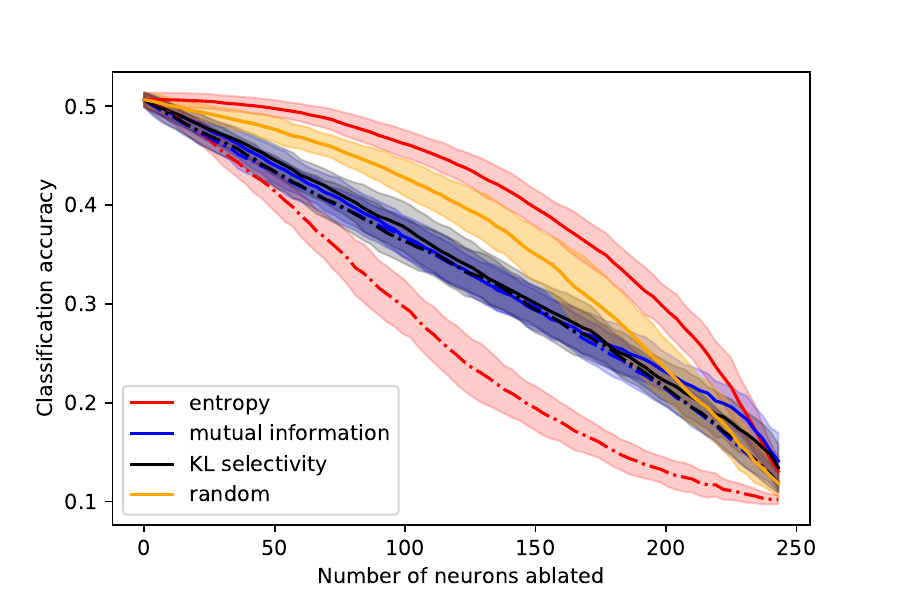}
        \includegraphics[width=0.24\textwidth,trim=0.5cm 0 1.2cm 0.5cm,clip]
        {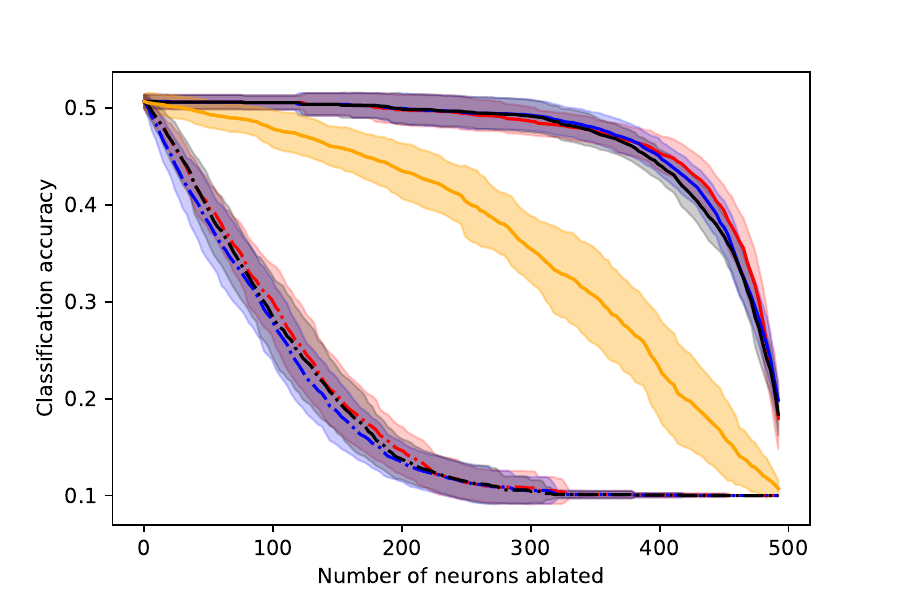} 
        \includegraphics[width=0.24\textwidth,trim=0.5cm 0 1.2cm 0.5cm,clip]
        {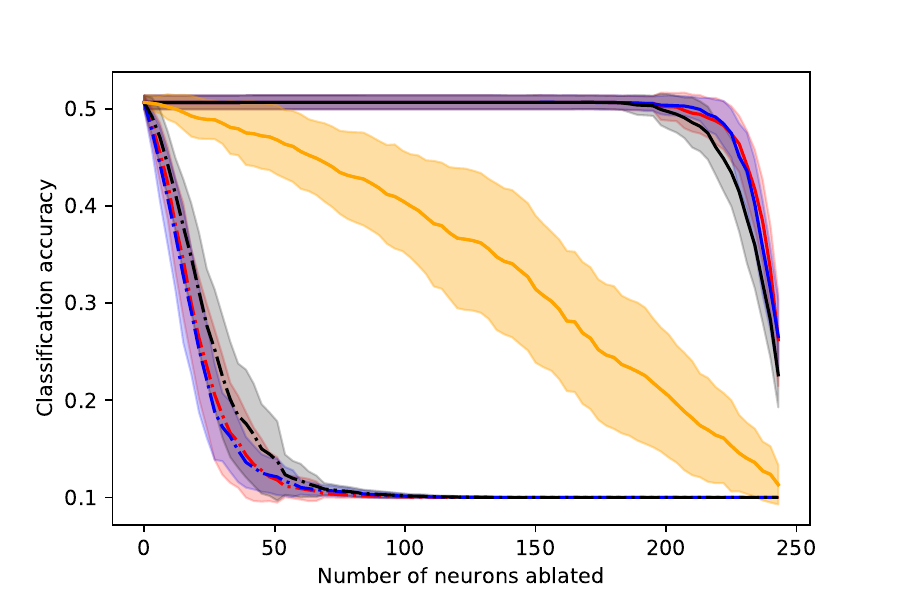}
        \includegraphics[width=0.24\textwidth,trim=0.5cm 0 1.2cm 0.5cm,clip]
        {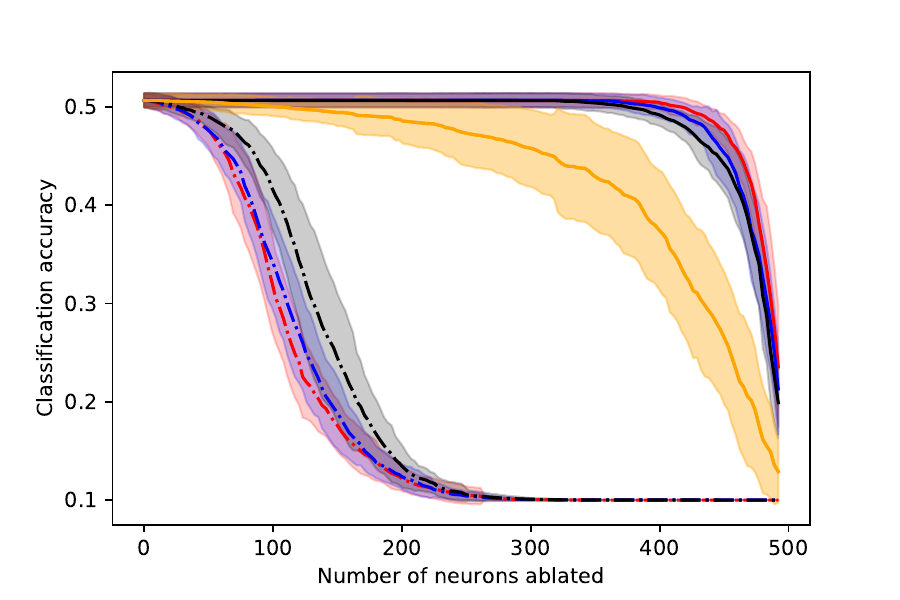}
    }
    \subfigure[Funnel CIFAR-10, ReLU, $L_2$ regularization, Layers 1 (left) to 4 (right)\label{fig:layerpruningCIFAR:Funnel} \label{fig:layerpruning:CIFAR}]
    {
        \includegraphics[width=0.24\textwidth,trim=0.5cm 0 1.2cm 0.5cm,clip]
        {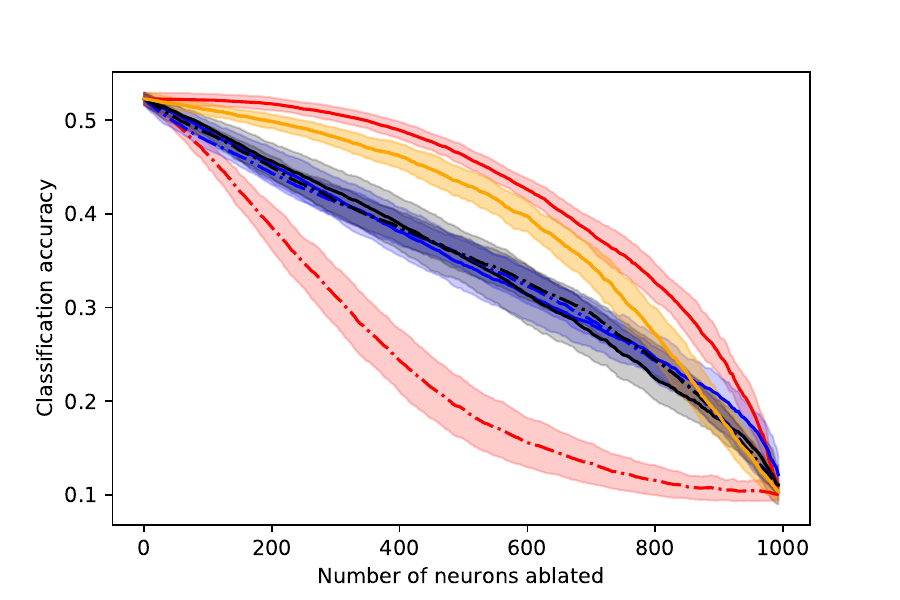}
        \includegraphics[width=0.24\textwidth,trim=0.5cm 0 1.2cm 0.5cm,clip]
        {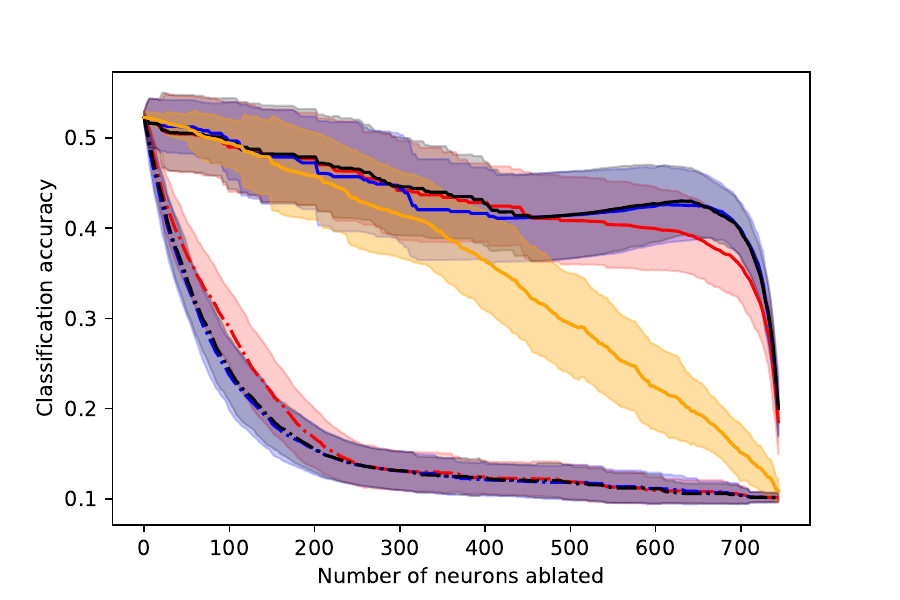} 
        \includegraphics[width=0.24\textwidth,trim=0.5cm 0 1.2cm 0.5cm,clip]
        {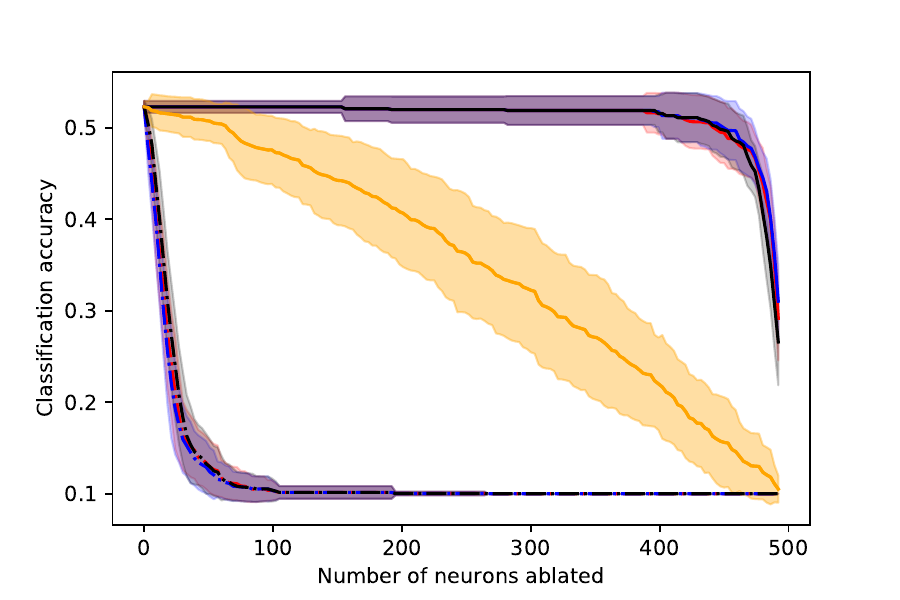}
        \includegraphics[width=0.24\textwidth,trim=0.5cm 0 1.2cm 0.5cm,clip]
        {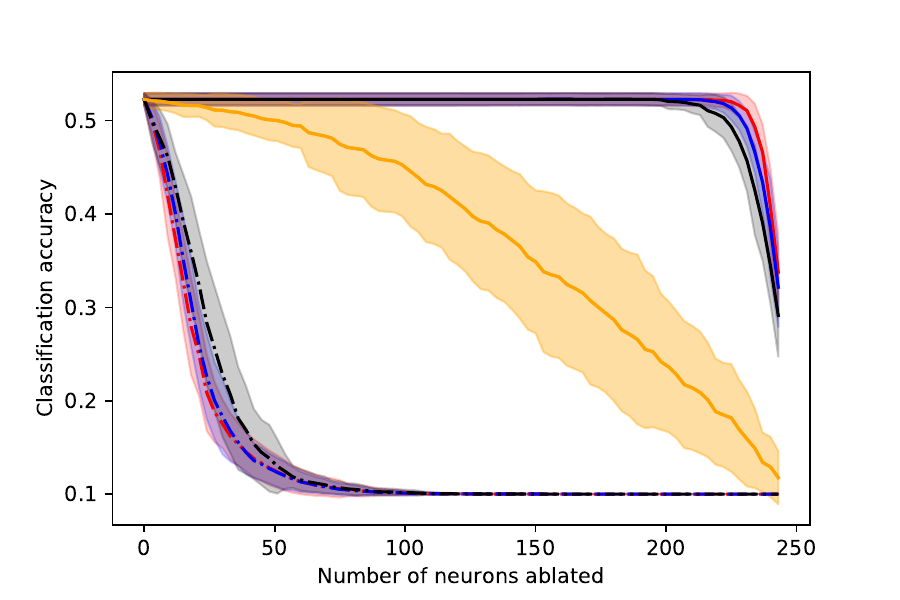}
    }
    \subfigure[Funnel CIFAR-10, ReLU, Dropout, Layers 1 (left) to 4 (right) \label{fig:layerpruningCIFAR:Dropout}]
    {
        \includegraphics[width=0.24\textwidth,trim=0.4cm 0 1.2cm 0.5cm,clip]
        {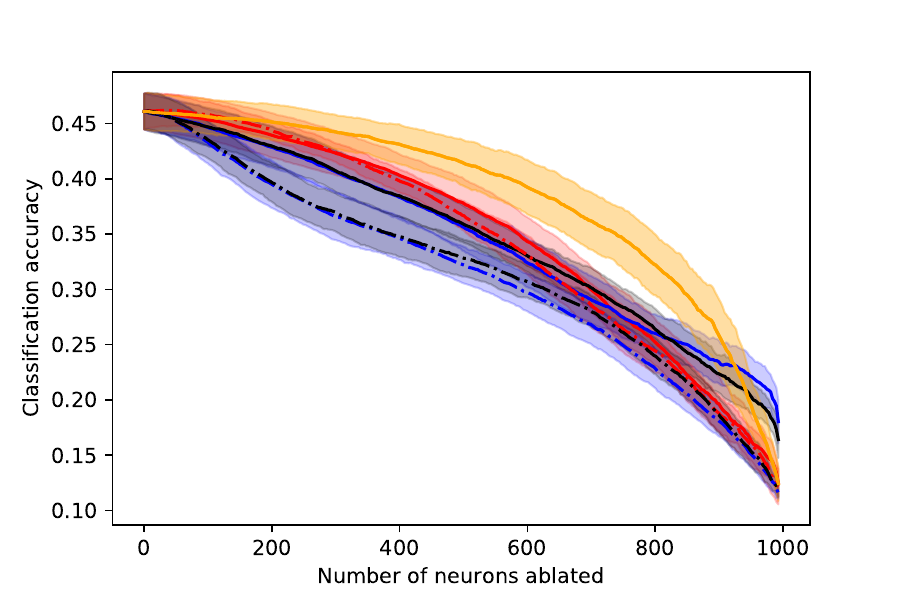}
        \includegraphics[width=0.24\textwidth,trim=0.4cm 0 1.2cm 0.5cm,clip]
        {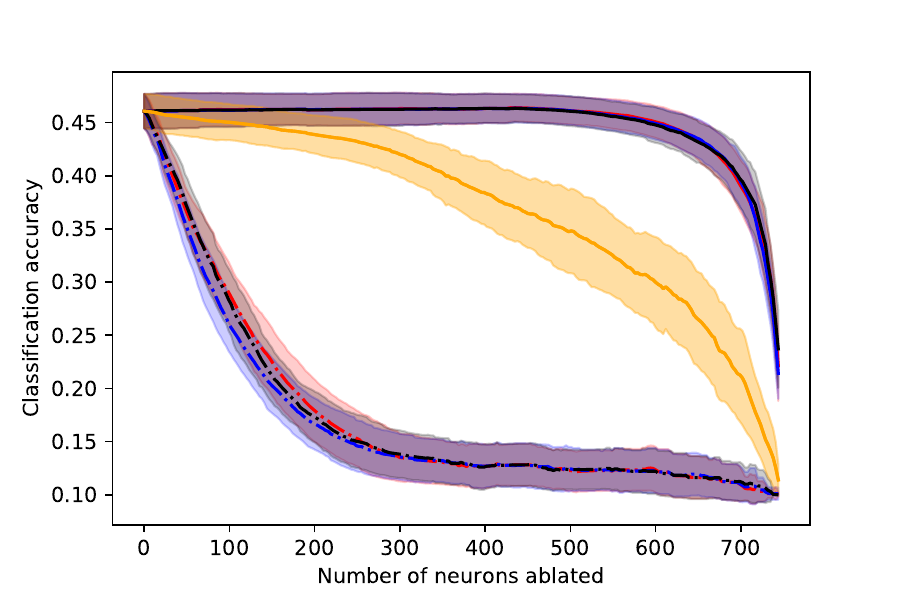} 
        \includegraphics[width=0.24\textwidth,trim=0.4cm 0 1.2cm 0.5cm,clip]
        {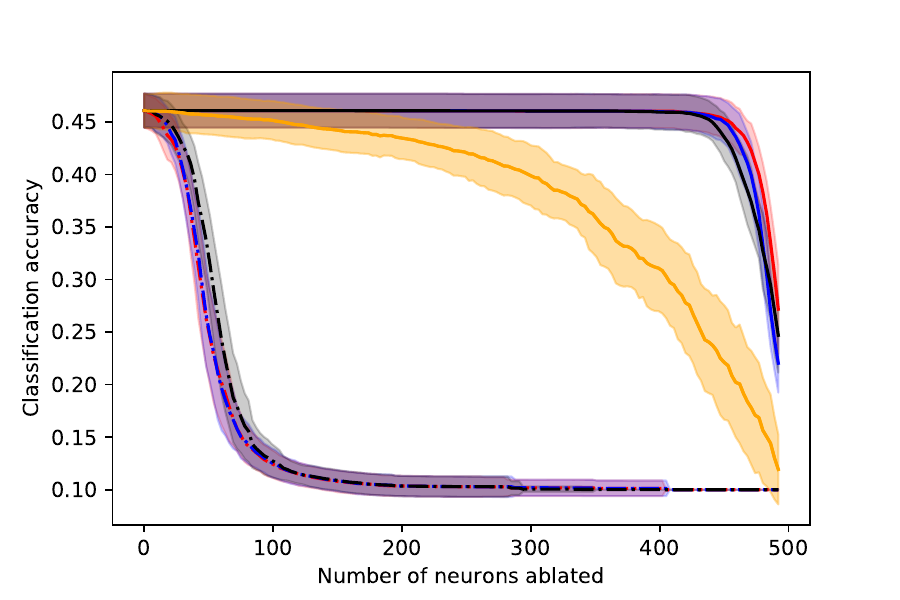}
        \includegraphics[width=0.24\textwidth,trim=0.4cm 0 1.2cm 0.5cm,clip]
        {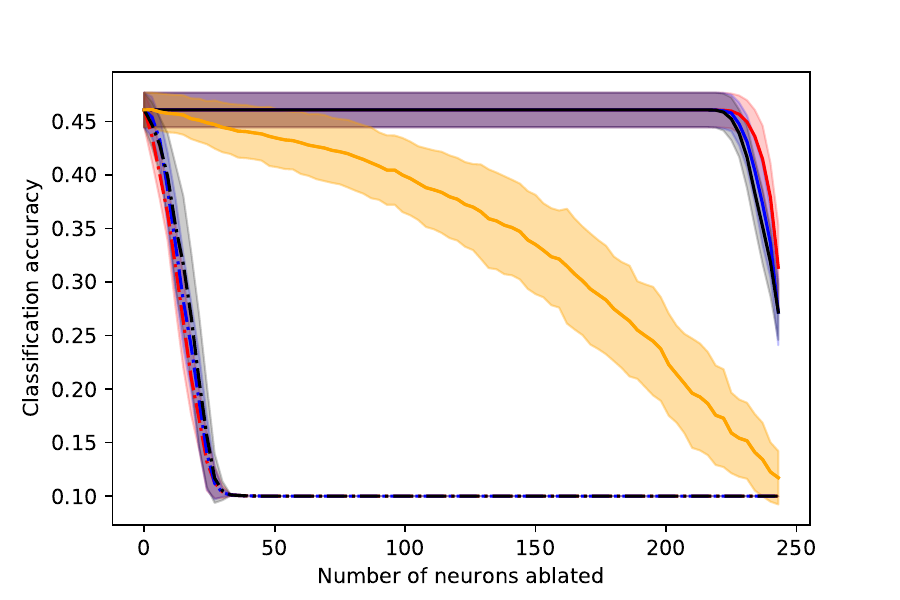}
    }
    \caption{Cumulative ablation of neurons in individual layers of NNs trained on CIFAR-10 confirm that entropy, mutual information, and KL selectivity are indicators for good classification performance in the second to forth layers of the NNs.}
    \label{fig:layerpruningCIFAR}
\end{center}
\end{figure*}

We finally summarize our results for NNs trained on CIFAR-10: As shown in Fig.~\ref{fig:layerpruningCIFAR}, in general, entropy, mutual information, and KL selectivity appear to be positively correlated with classification performance in the sense that ablating neurons with small (large) such values causes small (large) drops in classification accuracy. We now investigate the connection between the proposed candidate importance measures and classification performance as a function of layer number and training procedure in more detail and formulate hypotheses about the underlying effects.

Specifically, we first consider the ablation of neurons from the first hidden layer of NNs trained with $L_2$ regularization (Fig.~\ref{fig:layerpruningCIFAR:l2} and~\ref{fig:layerpruningCIFAR:Funnel}). There, one can see that ablating neurons from the first hidden layer with low mutual information or KL selectivity values does not perform better than ablating neurons with high such values, and that even random cumulative ablation performs better. Furthermore, it appears as if information-ordered cumulative ablation leads to an almost linear performance degradation. Such a situation can appear if the information is spread evenly over individual neurons, and the interplay between several such neurons is necessary to extract class information; in other words, such a situation could suggest high synergy. Indeed, the mutual information and KL selectivity values in the first layer are much smaller for NNs trained on CIFAR-10 than in those trained on MNIST or FashionMNIST (cf.~Fig.~\ref{fig:distribution}). Interestingly, entropy appears to be positively correlated with classification performance, at least for NNs trained with $L_2$ regularization. For the Funnel CIFAR-10 NN trained with Dropout (Fig.~\ref{fig:layerpruningCIFAR:Dropout}), random ablation outperforms any information-ordered cumulative ablation, which indicates that neurons with intermediate candidate importance values are contributing significantly to classification performance. While we believe that part of this effect is spurious (due to the narrow distributions of the considered values, cf.~Fig.~\ref{fig:distribution:CIFAR}), we also believe that Dropout is a major contributing factor: A NN trained with Dropout can be assumed to rely less on synergistic and unique information, but rather encourages redundancy. How this fact can help to explain the observed phenomenon will be the subject of future investigation.

Regarding the second hidden layer, the ablation analysis for the CIFAR-10 NN trained with with $L_2$ regularization in Fig.~\ref{fig:layerpruningCIFAR:l2} indicates that there are approximately 250 active neurons that contain all available information. Ablating top-ranked neurons immediately leads to a performance drop, which indicates that the information-theoretic candidate measures are all positively correlated with classification accuracy. For the Funnel CIFAR-10, $L_2$ regularization leads to peculiar results: Ablating the top-ranked 200 neurons almost disables the NN to perform classification. Ablating bottom-ranked neurons, however, also has negative effects on classification performance. Interestingly, a distributional analysis of the entropy values in the second hidden layer (such as the one in Fig.~\ref{fig:distribution:CIFAR} for the other CIFAR-10 NN) reveals that more than half of the neurons have an entropy equal to zero (thus, also their mutual information and KL selectivity values are zero). The fact that ablating these neurons has a negative effect indicates that these neurons are not inactive, but active all the time; see Example~\ref{ex:entropy}. This indicates that here the different activation values, rather than the fact whether they are active at all, convey (possibly synergistically) information relevant for classification, which is not captured by our candidate neuron importance measures; at least not with the used coarse quantization. Peculiar is also the fact that ablating the neurons with mutual information or KL selectivity ranks 500 through 600 \emph{improves} classification performance. We believe that also here synergistic effects could play a role: After initially ablating the 400 top-ranked neurons, the NN may be left with conflicting information that is clarified after continued ablation. While this hypothesis seems far-fetched, it is partly supported by Fig.~\ref{fig:layerpruningCIFAR:Dropout}, which shows that the same Funnel CIFAR-10 NN trained with Dropout does not show this behavior when ablating neurons from the second hidden layer.

The results for ablating neurons in the third and fourth hidden layer for the CIFAR-10 NN trained with $L_2$ regularization in Fig.~\ref{fig:layerpruningCIFAR:l2} and~\ref{fig:layerpruningCIFAR:Funnel} finally suggest that many neurons in these layers are inactive, which is again confirmed by the distribution of entropy values in Fig.~\ref{fig:distribution:CIFAR}. In the third hidden layer, there appear to be approximately 70 active neurons, characterized by positive mutual information and/or KL selectivity values. Ablating these 70 neurons completely disables the NN to perform classification; ablating all but these 70 neurons has negligible effect on classification performance. In the fourth layer of the same NN, approx. the 250 (CIFAR-10) or 70 (Funnel CIFAR-10) neurons with the highest information-theoretic quantities have to be ablated to strip the NN of its classification capabilities. In contrast, 100 (CIFAR-10) or 30 (Funnel CIFAR-10) of those neurons suffice to achieve full classification performance, indicating a certain degree of redundancy in the last hidden layer. This redundancy is higher in the 250-500-250-500 NN than in the 1000-750-500-250 NN, which is easily explained by the bottleneck structure of the former. The results for the Funnel CIFAR-10 NN trained with Dropout are similar: There, the third hidden layer displays some redundancy (the approx. 50 top-ranked neurons are sufficient for high classification performance, but up to approx. 150 top-ranked neurons need to be removed to reduce classification performance to chance), while the fourth hidden layer contains only approx. 25 active neurons (see Fig.~\ref{fig:layerpruningCIFAR:Dropout}).

\section{Conclusion, Impact, \& Limitations}\label{sec:discussion}

By considering validation data and neuron outputs as realizations of RVs, we defined and calculated information-theoretic quantities to measure the importance of individual neurons for classification. Using  cumulative ablation analyses, we arrived at the following main findings:
\begin{itemize}
 \item The distribution of the proposed quantities changes from layer to layer, with deeper layers in general having higher values (especially mutual information), cf.~\cite{Deepmind_Cats,Yu_CriticalPaths}. It is therefore ill-advised to compare neurons of different layers w.r.t.\ these measures (cf.~Sections~\ref{subsec:layerdistribution} and~\ref{subsec:pruning}).
 \item Cumulative ablation based on ordering \emph{individual} neurons using information-theoretic candidate measures allows to formulate hypotheses regarding the interplay of neurons of the \emph{whole} layer; e.g., cumulative ablation reveals redundancy, but may also hint at synergistic effects.
 \item In deeper layers, random ablation or ablation of low-ranking neurons has small effects on classification performance. This may be due to increased redundancy.
 \item The correlation between the considered quantities and classification performance depends on the activation function and regularization. For example, the correlation is strongly positive for NNs with ReLU activation functions trained with $L_2$ regularization and with Dropout. In contrast, the correlation is weak for NNs with sigmoid activation functions (Fig~\ref{fig:layerpruningMNIST:sigmoid}) and for NNs trained with a combination of Dropout and batch normalization (see \cite[Appendices~C~\&~D]{Amjad_Understanding_arXiv}). This may be attributed to our comparably simple and coarse quantization scheme.
 \item The correlation between the considered quantities and neuron importance depends on the depth and structure of the considered layer, and on the NN architecture as a whole. For example, wide layers may have many inactive neurons (leading to stronger correlation, Fig.~\ref{fig:layerpruningCIFAR}), while deep layers may have many redundant neurons (leading to weaker correlation, Fig.~\ref{fig:layerpruningMNIST}).
\end{itemize}

We anticipate impact of our results in three different areas: First and foremost, our analysis provides insight into the inner workings of NNs. Such insights can be used to develop posthoc model interpretability approaches that are required especially in cases where NNs shall be applied in safety- or privacy-critical applications, or in cases where the suggestions of the NN-based system have ethical implications. As a concrete example, we would like to point at the recent discussion of whether successful NNs are compiled of highly class-selective or informative neurons~\cite{Deepmind_Cats,Zhou_RevisitingCNNAblation,Meyes_Ablation,Ukita_Orientation}. Our general conclusion that mutual information and KL selectivity of a neuron output are correlated with this neuron's importance in the classification task seems to contradict the opposite claims in~\cite{Deepmind_Cats}. This contradiction is resolved by combining, e.g., the observations from Figs.~\ref{fig:distribution} and~\ref{fig:layerpruningMNIST}: The mutual information of neurons in the second layer of the NN trained on MNIST are large compared to those of the first layer; simultaneously, neurons in the second layer seem to be highly redundant and can be removed without affecting classification performance. Thus, ablating neurons with large mutual information values affects classification performance less than ablating neurons with small mutual information values, leading to the negative correlation reported in~\cite{Deepmind_Cats}. We have argued, however, that the correlation between mutual information and impact on classification performance becomes positive if this correlation is evaluated layer-by-layer. Thus, the apparent conflict between our results and those in~\cite{Deepmind_Cats} is an instance of Simpson's paradox and can be resolved by recognizing that it is ill-advised to compare mutual information and (KL) selectivity values across different layers.

Second, this improved understanding about which properties of a neuron make it an important contributor to NN performance can lead to the design of loss functions and/or regularizers for NN training. For example, once it is understood that deeper layers are characterized by higher mutual information with the class variable, this characteristic can be explicitly rewarded by designing appropriate loss function terms.

Third, we believe that these results can influence post-training optimization of NNs, such as weight compression and pruning \cite{Kuzmin19}, weight quantization \cite{wang20, choukroun19}, neuron pruning~\cite{he2014reshaping,Li_CNNPruning,Molchanov_CNNPruning}, or merging of neurons that behave similarly~\cite{srinivas2015data,Mariet_Divnet}. For example, our conclusion from Figs.~\ref{fig:wholeNNpruning}~\ref{fig:layerpruningMNIST},~\ref{fig:layerpruningFashionMNIST}, and~\ref{fig:layerpruningCIFAR} is that, if the neurons to be pruned are selected based on information-theoretic quantities, pruning has to be performed layer by layer rather than in the entire NN at once. We believe this observation may be true for other quantities as well; e.g., for $L_2$-based weight pruning, the distribution of weight magnitudes may differ from layer to layer. We further observed that our information-theoretic quantities not only differ greatly between layers (see Fig.~\ref{fig:distribution}) but also have different meanings (see Section~\ref{sec:impfunc} and \cite[Appendix~F]{Amjad_Understanding_arXiv}). This suggests that it may be useful to employ different quantities when pruning different layers. Finally, our discussion in Section~\ref{subsec:layerpruning:MNIST} indicates that deeper layers have more redundancy and hence can be more severely pruned without impacting the performance significantly. To operationalize this, one requires measures of neuron redundancy beyond a simple comparison of in- and outgoing weights, such as in~\cite{he2014reshaping}.

We finally point at two limitations of our study and how these can be lifted. On the one hand, our experiments are currently restricted to fully-connected NNs. Furthermore, we have observed that results for NNs with sigmoid activation functions or trained using batch normalization are inconclusive, showing at most weak correlation between classification accuracy and the proposed candidate measures. The question whether our qualitative claims hold more generally (e.g., for underparameterized NNs that underfit, NNs with bottleneck layers, particularly wide layers, slim but deep NNs, or for NNs trained with different regularizers, etc.), shall be answered in future work. Specifically, we believe that finer quantization may shed light on the effects of batch normalization that will help understanding the resulting NNs better. The analysis of different NN classes such as convolutional NNs, recurrent NNs, or residual NNs, will require more careful tailoring of our current approach. For example, a naive way to compute the proposed information-theoretic quantities for convolutional layers of convolutional NNs is to consider each 'pixel' of a latent activation map separately. However, as the 'pixels' in the same output channel of the activation feature map have high correlation, we believe that one should define quantities that represent the information content of each output channel instead. Developing a meaningful approach to calculating information content for such high-dimensional variables (such that the information content is also related to the task in an interesting manner) is, however, an elaborate research problem in its own right (cf.~\cite{Belghazi_MINE}).

On the other hand, our proposed quantities are based on \emph{individual} neurons, ignoring potential correlations between different neurons of the same layer. Our work is thus at one extreme end; the recently proposed information bottleneck theory of NNs, which computes a single mutual information value for an entire layer~\cite{Tishby_BlackBox} lies at the opposite end. Neither is sufficient to characterize NNs: We point at shortcomings of our approach in Examples~\ref{ex:entropy} through~\ref{ex:misynergy} and refer the reader to~\cite{Saxe_IBTheory,Amjad_HowNotTo,Goldfeld_Estimating} for shortcomings of the information bottleneck theory of NNs. We believe that a full information-theoretic characterization of NNs requires additional mathematical quantities, computed also for subsets of neurons of a given layer. For example, the shortcomings highlighted in Examples~\ref{ex:klred} and~\ref{ex:mired} can possibly be accounted for by introducing quantities that take the redundancy of a layer into account, such as those proposed by~\cite{srinivas2015data,babaeizadeh2016noiseout}. For Example~\ref{ex:klred}, another option is to replace the KL selectivity of a neuron by its \emph{specific information spectrum}, $\{\kld{P_{T_j^{(i)}|Y=y}}{P_{T_j^{(i)}}}\}_{y\in\setnotation{C}}$. The resulting spectra, evaluated for every neuron in a given layer, allow selecting a subset of neurons such that each class in $\setnotation{C}$ is represented.  More generally, all examples can be treated by investigating how the mutual information between a complete layer and the class variable splits into redundant, unique, and synergistic parts. Neurons that contain only redundant information shall be assigned little importance; in deeper layers, unique information may be given higher value than synergistic information, whereas the contrary may be true for shallow layers. This line of thinking suggests that PID~\cite{Williams_PID,Rauh_PID} can shed more light on the behavior of NNs. For example, the authors of~\cite{Tax_PID} discover two distinct phases during training a NN with a single hidden layer, characterized by large amounts of redundant and unique information, respectively. Computing the PID of a wide layer is prohibitive in terms of computational complexity and suffers from the curse of dimensionality. An approximate PID, based on bounds and approximations, may be a computationally feasible way of validating the claims regarding synergy and redundancy made at the hand of information-ordered cumulative ablation. Future work shall be devoted to this analysis.

\bibliographystyle{IEEEtran}
\bibliography{IEEEabrv,references}

\begin{IEEEbiography}[{\includegraphics[width=1in,height=1.25in,clip,keepaspectratio]{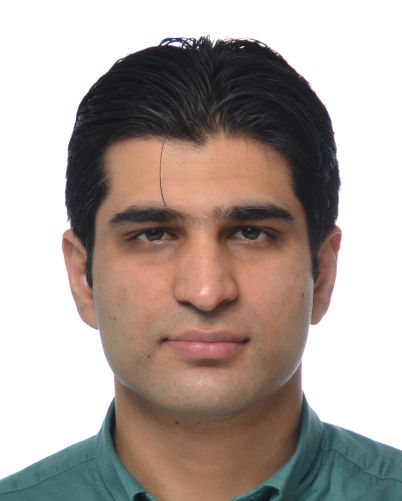}}]{Rana Ali Amjad}
(S'13, M'19) was born in Sahiwal, Pakistan, in 1989. He received his bachelors degree in electrical engineering (with highest distinction) from University of Engineering and Technology, Lahore, Pakistan, in 2011. He completed his masters degree in communication engineering (with highest distinction) and his PhD in electrical engineering and information technology from Technical University of Munich, Germany, in 2013 and 2019 respectively. During his PhD he worked on information theory, graphical models and algorithms for problems in telecommunications and machine learning. 

In July 2019 he joined Qualcomm AI research in Amsterdam, Netherlands as a deep learning research scientist, where his main topics of research were efficient deep learning inference  and neural augmentation of classical machine learning algorithms for wireless communication applications. Since April 2021 he is an applied research scientist at Amazon A9 Search organization in Palo Alto, USA where he conducts research on efficient deep learning systems that can be trained and deployed at scale for a multitude of applications. His research interests cover information theory, deep learning, causal inference and reasoning, and communication theory.
\end{IEEEbiography}

\begin{IEEEbiography}[{\includegraphics[width=1in,height=1.25in,clip,keepaspectratio]{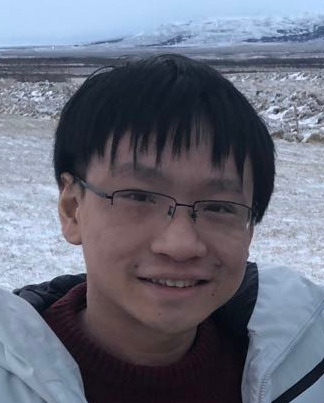}}]{Kairen Liu} was born in Shanghai, China, in 1992. He received the Bachelor's degree in Mechatronics from Tongji University, Shanghai, China, In 2012. He completed his Master’s degree in Electrical Engineering from Technical University of Munich, Germany, in 2017. 

Since 2017 he has been working in the telecommunication industry involving business data analysis, and Digital transformation. His academic interests cover information theory, machine learning and other coding related topics.
\end{IEEEbiography}

\begin{IEEEbiography}[{\includegraphics[width=1in,height=1.25in,clip,keepaspectratio]{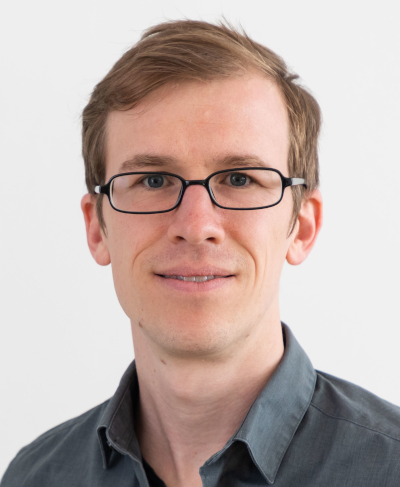}}]{Bernhard C. Geiger}
(S'07, M'14, SM'19) received the Dipl.-Ing. degree in electrical engineering (with distinction) and the Dr. techn. degree in electrical and information engineering (with distinction) from Graz University of Technology, Austria, in 2009 and 2014, respectively.

In 2009 he joined the Signal Processing and Speech Communication Laboratory, Graz University of Technology, as a Project Assistant and took a position as a Research and Teaching Associate at the same lab in 2010. He was a Senior Scientist and Erwin Schr\"odinger Fellow at the Institute for Communications Engineering, Technical University of Munich, Germany from 2014 to 2017 and a postdoctoral researcher at the Signal Processing and Speech Communication Laboratory, Graz University of Technology, Austria from 2017 to 2018. He is currently a Senior Researcher at Know-Center GmbH, Graz, Austria, where he also leads the Machine Learning Group within the Knowledge Discovery Area. His research interests cover information theory for machine learning, theory-assisted machine learning, and information-theoretic model reduction for Markov chains and hidden Markov models.
\end{IEEEbiography}

\end{document}